\documentclass{article}
\usepackage{ijcai23}

\usepackage{times}
\usepackage{soul}
\usepackage{url}
\usepackage[hidelinks]{hyperref}
\usepackage[utf8]{inputenc}
\usepackage[small]{caption}
\usepackage{epsfig}
\usepackage{graphicx}
\usepackage{amsmath}
\usepackage{amsthm}
\usepackage{amssymb}
\usepackage{booktabs}
\usepackage{algorithm}
\usepackage{algorithmic}
\usepackage{multicol}
\usepackage{multirow}
\usepackage{listings}
\usepackage[dvipsnames]{xcolor}
\usepackage{tcolorbox}
\usepackage[switch]{lineno}

\definecolor{font_color}{rgb}{0.5,0.5,0.5}
\definecolor{background_color}{rgb}{0.92,0.95,0.95}
\lstdefinestyle{mystyle}{
    backgroundcolor=\color{background_color},   
    commentstyle=\color{font_color},
    keywordstyle=\color{font_color},
    numberstyle=\tiny\color{font_color},
    stringstyle=\color{font_color},
    basicstyle=\ttfamily\footnotesize,
    breakatwhitespace=false,         
    breaklines=true,                 
    captionpos=b,                    
    keepspaces=true,              
    numbersep=5pt,                  
    showspaces=false,                
    showstringspaces=false,
    showtabs=false,
}
\title{Contour-based Interactive Segmentation}

\author{
Polina Popenova$^1$
\and
Danil Galeev$^1$\and
Anna Vorontsova$^{1}$\And
Anton Konushin$^1$\\
\affiliations
$^1$Samsung Research\\
\emails
\{p.popenova, d.galeev, a.vorontsova, a.konushin\}@samsung.com
}

\begin{document}

\maketitle

\def\dset{UserContours}
\def\dsetgroup{UserContours-G}
\def\dsetsize{2000}
\def\dsetgroupsize{50}

\begin{abstract}

Recent advances in interactive segmentation (IS) allow speeding up and simplifying image editing and labeling greatly. The majority of modern IS approaches accept user input in the form of clicks. However, using clicks may require too many user interactions, especially when selecting small objects, minor parts of an object, or a group of objects of the same type. In this paper, we consider such a natural form of user interaction as a loose contour, and introduce a contour-based IS method. We evaluate the proposed method on the standard segmentation benchmarks, our novel \dset{} dataset, and its subset \dsetgroup{} containing difficult segmentation cases. Through experiments, we demonstrate that a single contour provides the same accuracy as multiple clicks, thus reducing the required amount of user interactions.
\end{abstract}

\section{Introduction}

IS aims to segment an arbitrary object in an image according to a user request. IS has numerous applications in image editing and labeling: it can significantly speed up labeling images with per-pixel masks and ease the burden of annotating large-scale databases~\cite{acuna2018efficient,agustsson2019interactive,benenson2019large}. In graphical editors, IS might allow users selecting objects of interest to manipulate them.

\begin{figure}[h!]
    \centering
    \includegraphics[width=0.95\linewidth]{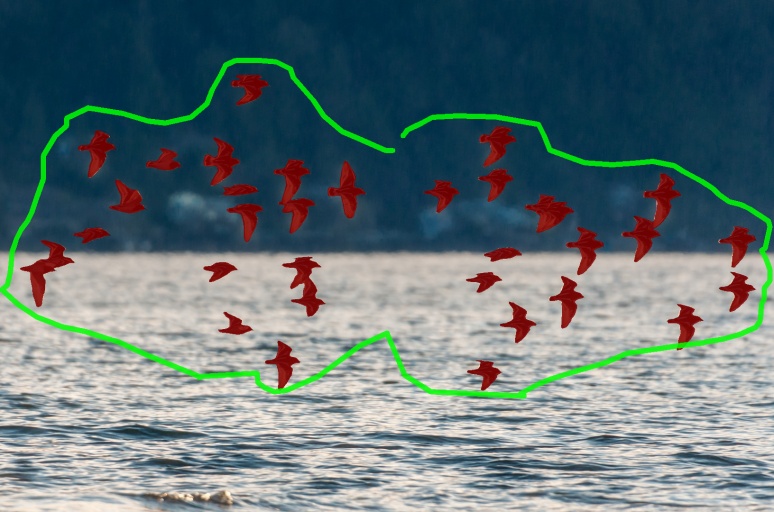}
    \caption{An example of an image where selecting objects with contours is much more efficient than with clicks. It takes about 5 seconds to select a flock of birds with a single contour, whereas it takes about 40 seconds with clicks (clicking on \textit{each} bird in the flock!).}
    \label{fig:teaser}
\end{figure}

Recent IS works~\cite{hao2021edgeflow,jang2019interactive,sofiiuk2020f,sofiiuk2021reviving,xu2016deep} consider user input in the form of clicks. A simple and intuitive form of user interaction, clicks are not always the best option for object selection. For instance, it is difficult to click precisely on a tiny object on the small smartphone screen. Lately, sliding with a finger tends to replace tapping: Word Flow keyboard featuring shape writing was officially certified as the fastest smartphone keyboard, and many users prefer drawing a pattern to unlock the screen rather than entering a PIN-code with several clicks. Hence, we assume that switching from discrete to continuous input speeds up the interaction, and for smartphone users, it would be more convenient to draw a contour rather than to click for several times.

Accordingly, we focus on the \textbf{contour-based IS}. We address this task with a novel trainable method that segments an object of interest given a single contour.
Our method does not require manually annotated contours for training but makes use of conventional segmentation masks, so it can be trained on the standard segmentation datasets such as LVIS~\cite{gupta2019lvis}, COCO~\cite{lin2014microsoft}, OpenImages~\cite{kuznetsova2020open}, and SBD~\cite{hariharan2011semantic}. Our experiments show that a single contour allows achieving the same accuracy as 3-5 clicks on the standard benchmarks: GrabCut~\cite{rother2004grabcut}, Berkeley~\cite{martin2001database,mcguinness2010comparative}, and DAVIS~\cite{li2018interactive,perazzi2016benchmark}.
We also present \textbf{\dset}, a collection of \dsetsize{} images of common objects in their usual environment, annotated with real-user contours. Besides, we create the \textbf{\dsetgroup} dataset by selecting \dsetgroupsize{} images especially difficult to segment: those depicting small objects, overlapped objects, and groups of objects. We empirically prove that our contour-based approach has an even greater advantage on such challenging human-annotated data compared to the click-based approach.

Overall, our contribution can be summarized as follows:
\begin{itemize}
    \item To the best of our knowledge, we are the first to formulate the task of IS given a single contour;
    \item We adapt a state-of-the-art click-based model for contours, while not sacrificing its inference speed;
    \item We introduce a \dset{} dataset and a challenging \dsetgroup, manually labeled with contours;
    \item We develop an evaluation protocol that allows comparing the contour-based and click-based methods, and show that a single contour is equivalent to multiple clicks (up to 20!) in terms of segmentation accuracy.
\end{itemize}

\section{Related work}

IS aims at obtaining a mask of an object given an image and an additional user input. Early methods \cite{boykov2001interactive,grady2006random,gulshan2010geodesic,rother2004grabcut} tackle the task via minimizing a cost function defined on a graph over image pixels. 

\paragraph{Click-based methods.}
Xu et al.~\cite{xu2016deep} introduced a CNN-based method and a click simulation strategy for training click-based IS methods on the standard segmentation datasets without additional annotation. In \cite{li2018interactive,liew2017regional,liew2019multiseg,lin2020interactive}, network predictions are refined through attention. BRS~\cite{jang2019interactive} minimized a discrepancy between the predicted mask and the map of clicks after each click, while in~\cite{sofiiuk2020f,kontogianni2020continuous}, inference-time optimization was applied to higher network levels. Recent click-based  methods~\cite{hao2021edgeflow,jang2019interactive,sofiiuk2020f,sofiiuk2021reviving} show impressive accuracy, but still may require a lot of interactions. Among them, the best results are obtained via iterative click-based approaches~\cite{jang2019interactive,sofiiuk2020f,sofiiuk2021reviving} that leverage information about previous clicks. In such methods, model weights are updated after each user input, which increases the computational cost per click.

\paragraph{Alternative user inputs.} Alongside numerous click-based methods, other types of user input have been investigated. Strokes were widely employed as a guidance~\cite{andriluka2020efficient,bai2014error,batra2010icoseg,boykov2001interactive,freedman2005interactive,grady2006random,gueziri2017latency,gulshan2010geodesic,kim2008generative,lin2016scribblesup}; however, no comparison with click-based approaches was provided. Putting a stroke requires a lot of effort, and most stroke-based methods employed training-free techniques to imitate user inputs.
DEXTR~\cite{maninis2018deep} used extreme points: left, right, top, and bottom pixels of an object. In a recent work~\cite{ferrari2019scribbles}, strokes were combined with extreme points. However, placing extreme points in the right locations is non-trivial and definitely harder than clicking on an object, and the predictions cannot be corrected as well. Bounding boxes were used either for selecting large image areas~\cite{cheng2015densecut,rother2004grabcut,wu2014milcut,xu2017deep} or segmenting thin objects~\cite{liew2021tos}. The main drawbacks of bounding boxes are lack of specific object reference inside the selected area and no support for correcting predicted mask. However, it was shown that a model trained on bounding boxes could generalize to arbitrary closed curves~\cite{xu2017deep}. In~\cite{zhang2020interactive}, bounding boxes are combined with clicks giving more specific object guidance and facilitating corrections. PhraseClick \cite{ding2020phraseclick} combined clicks with text input to specify object attributes and reduce the number of clicks. 

We consider user input in the form of contours, and build a network capable of processing contours by slightly modifying a click-based IS network. This approach is proved to outperform click-based methods: particularly, we show that a single contour provides better results than several clicks.

\section{Contour-based IS Method}
\label{sec:method}

\begin{figure}[h!]
    \begin{center}
        \includegraphics[width=0.95\linewidth]{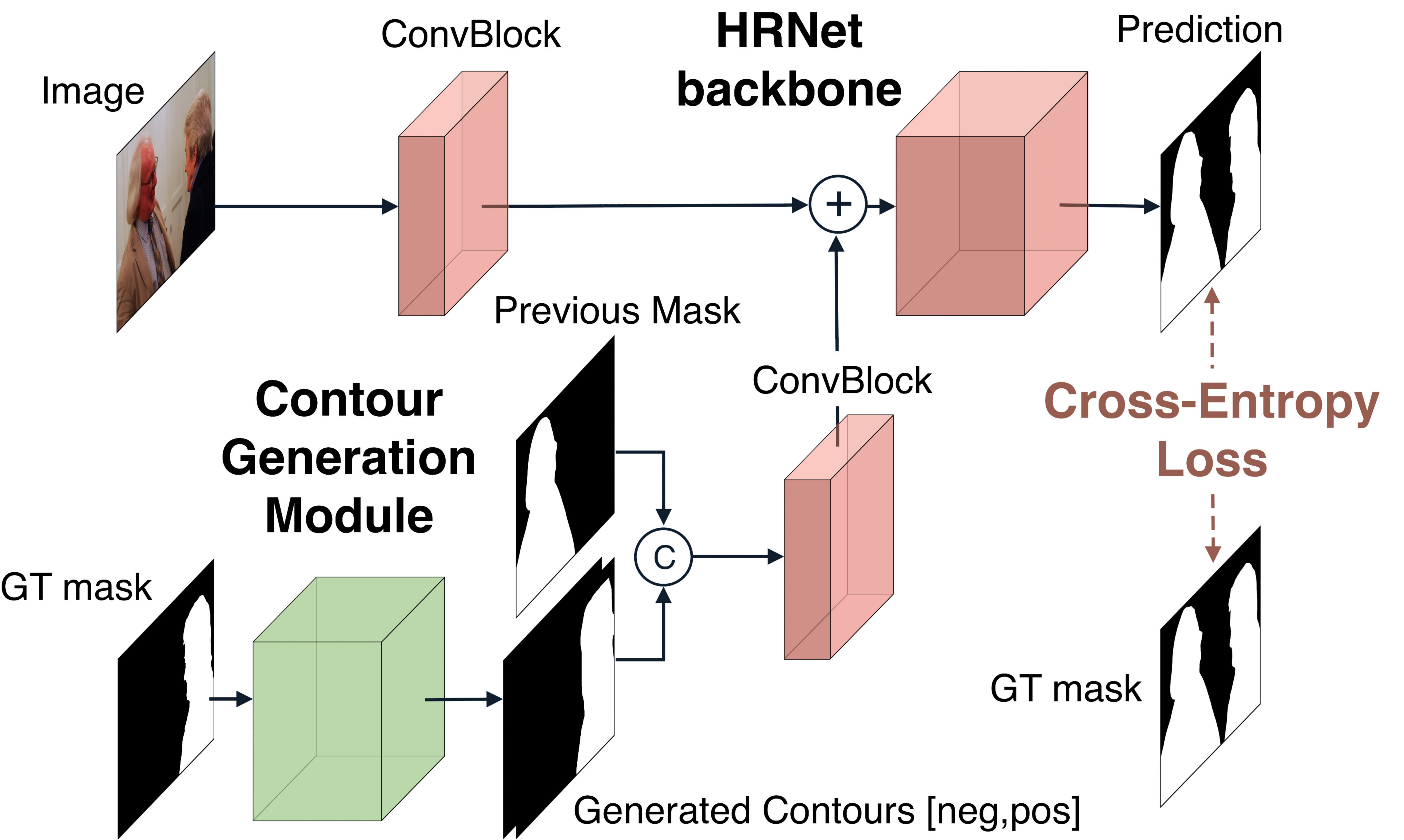}
    \end{center}
    \caption{The architecture of the proposed method. The contour generation module simulates user contours. The generated contours are encoded as binary masks, stacked with a mask from a previous interaction, and fed into the network via a novel interactive branch. The network is trained to minimize binary cross-entropy between a predicted and a ground truth mask.}
\label{fig:method}
\end{figure}

Our method is inherited from the state-of-the-art click-based RITM~\cite{sofiiuk2021reviving}, having an interaction generation module, a backbone, and an interactive branch (Fig.~\ref{fig:method}). 

\begin{table*}[h!]
\setlength{\tabcolsep}{1pt}{
\begin{center}
\resizebox{0.95\linewidth}{!}{
\begin{tabular}{ccccccccc}
    \small
    \multirow{3}{*}{Image} & \multirow{3}{*}{GT mask} & (1) & (2) & (3) & (4) & (5) & (6) & \multirow{3}{*}{\shortstack{Generated\\contour}} \\
    & & Dilation / & Largest & Elastic & Gaussian & Scale & Shift &  \\
    & & erosion & component & transform & Blur & & &  \\
    \includegraphics[width=60pt]{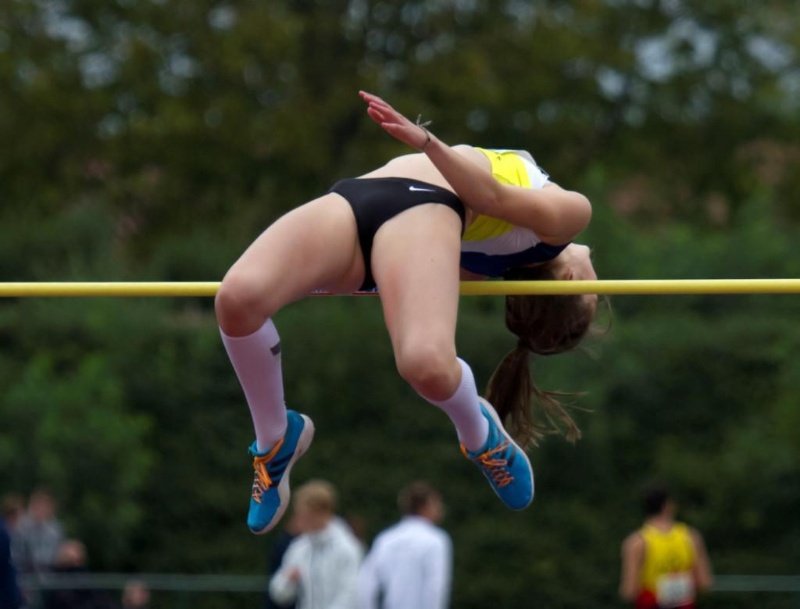} &
    \includegraphics[width=60pt]{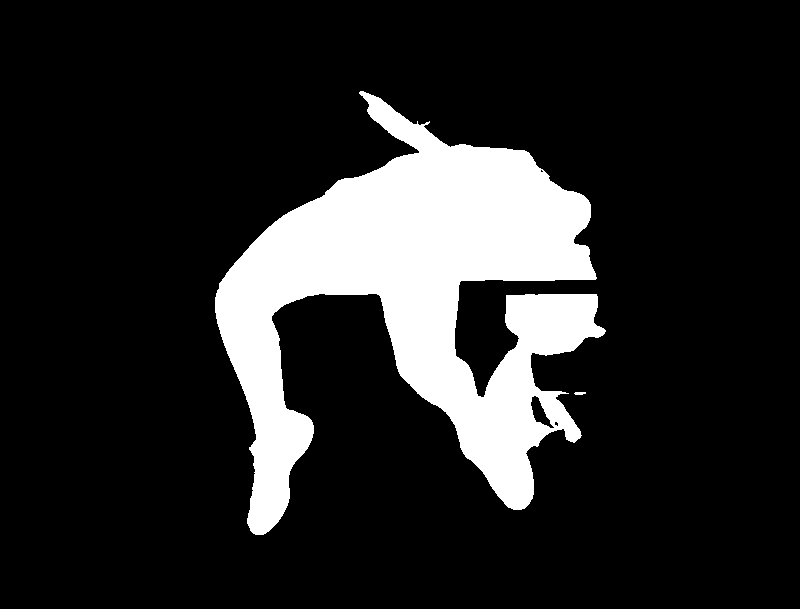} &
    \includegraphics[width=60pt]{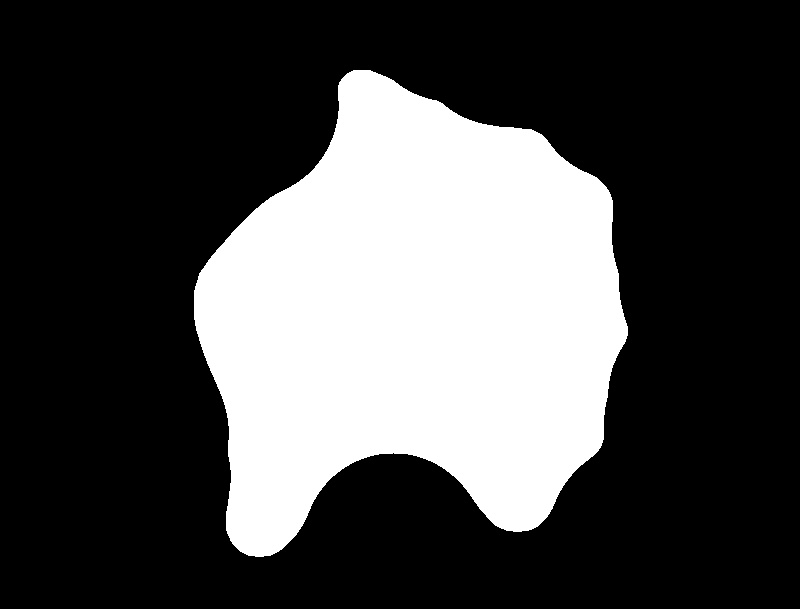} &
    \includegraphics[width=60pt]{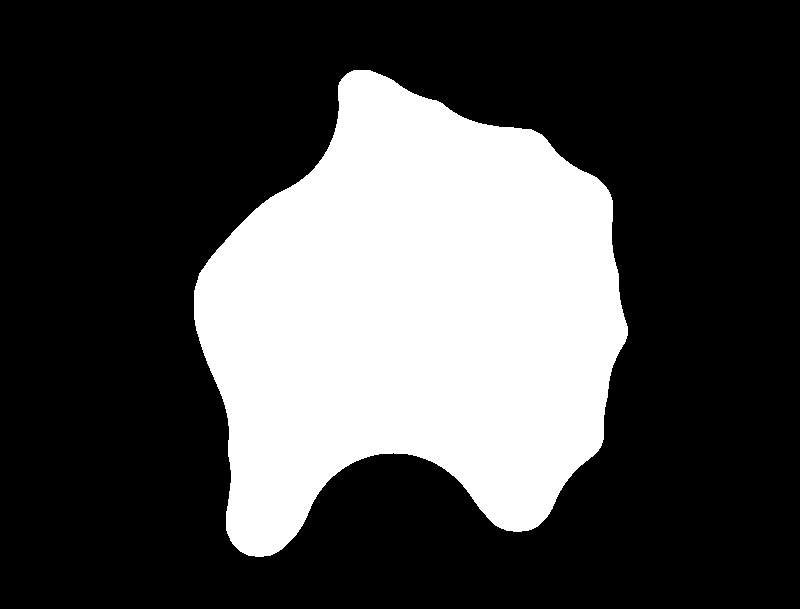} &
    \includegraphics[width=60pt]{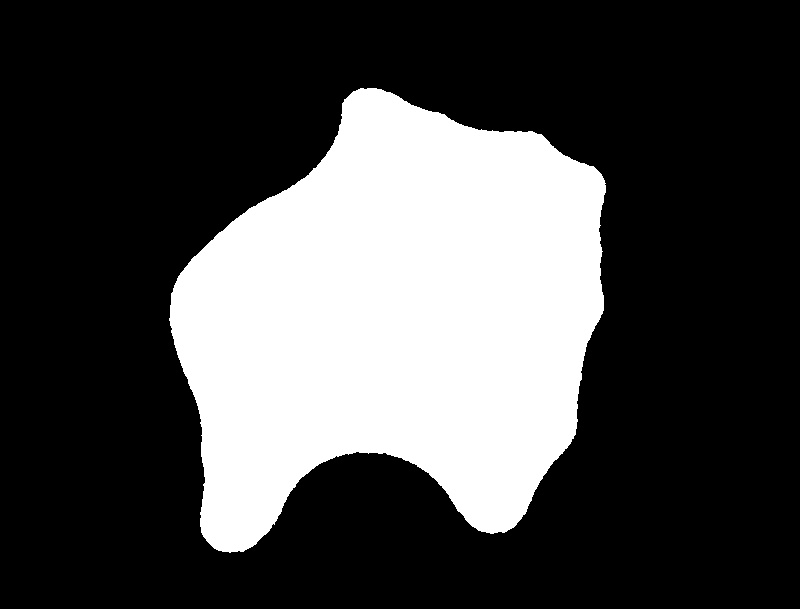} &
    \includegraphics[width=60pt]{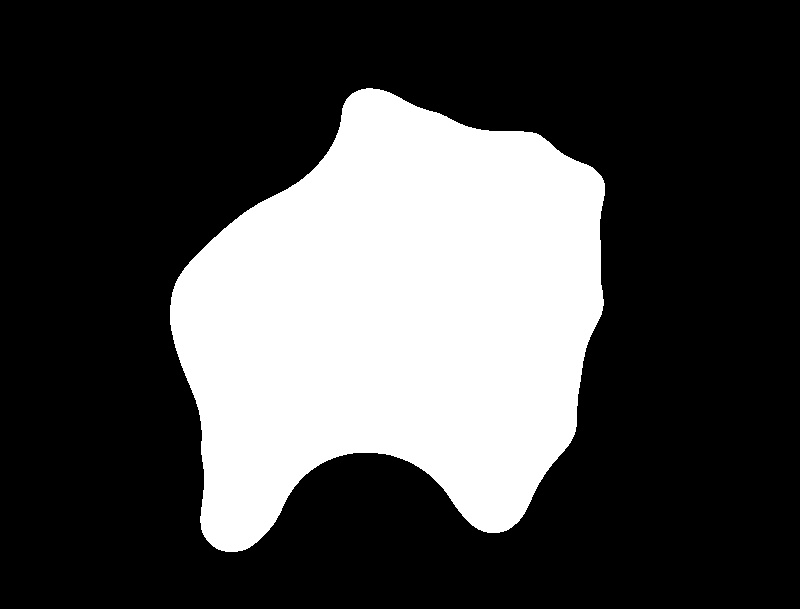} &
    \includegraphics[width=60pt]{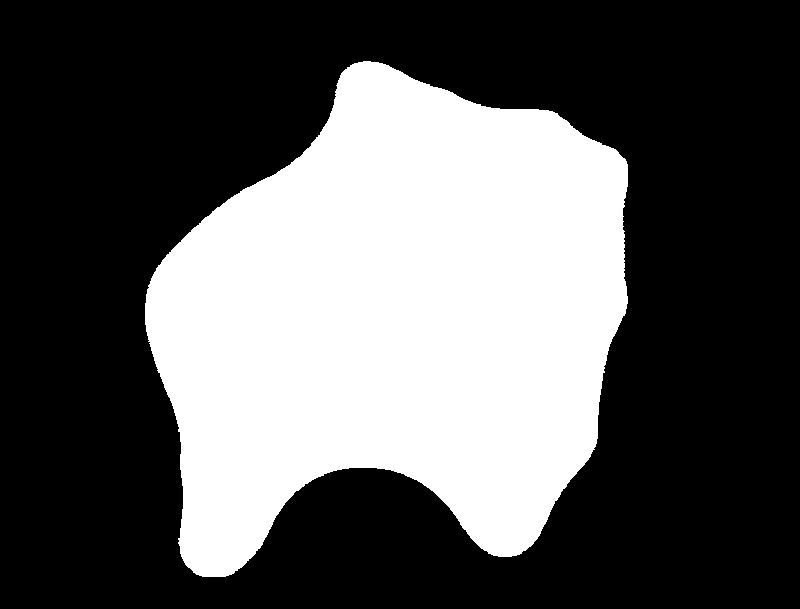} &
    \includegraphics[width=60pt]{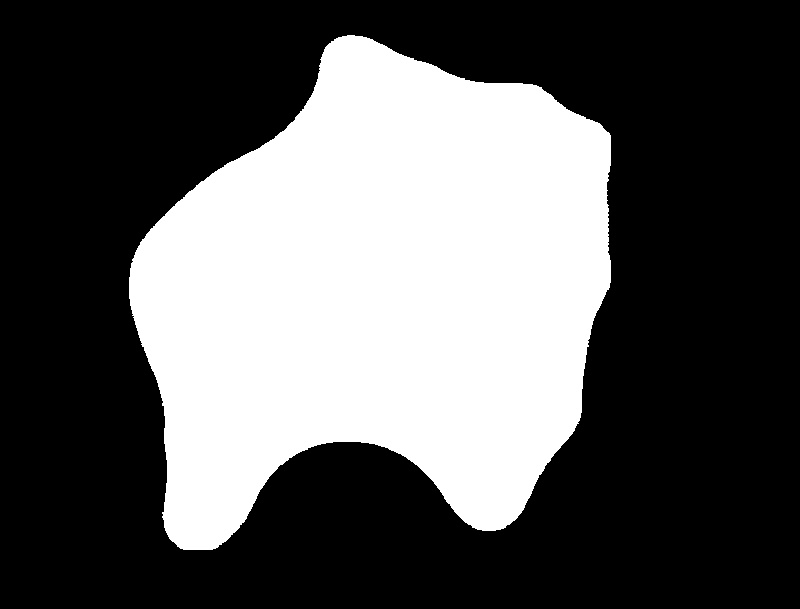} &
    \includegraphics[width=60pt]{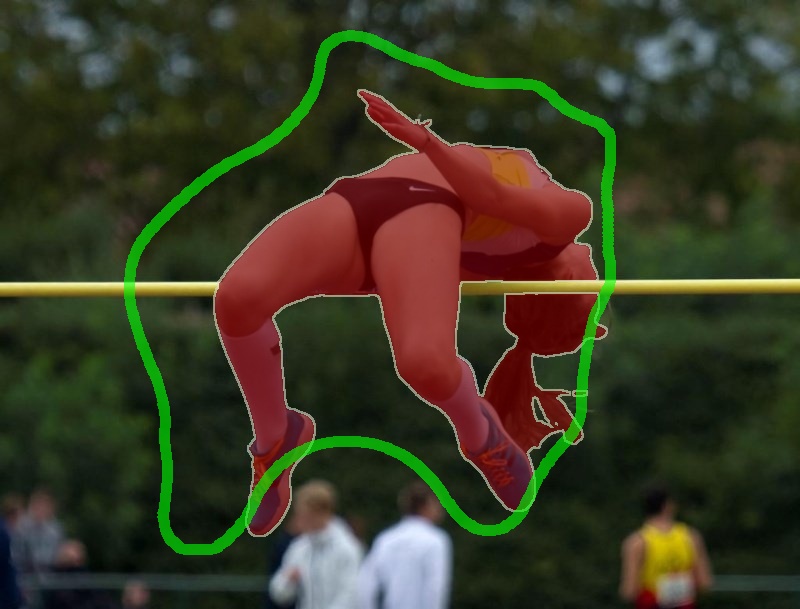} \\
    \includegraphics[width=60pt]{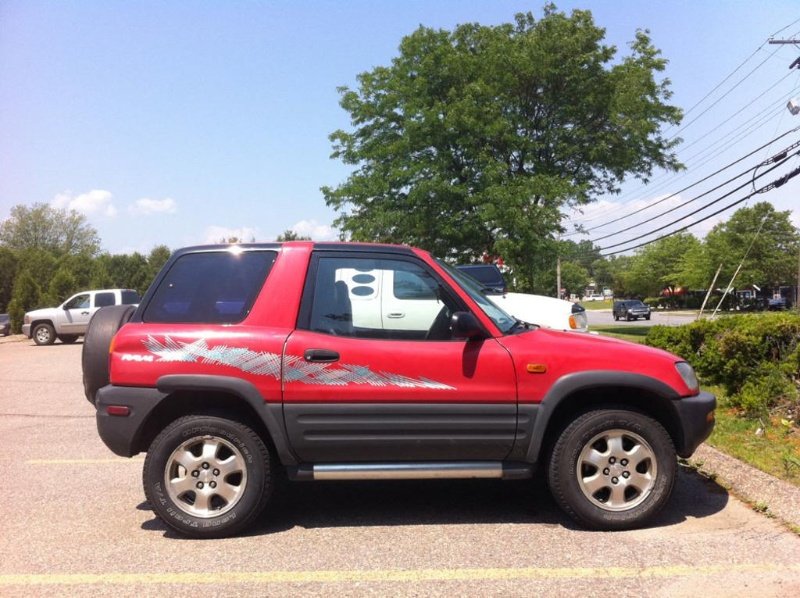} &
    \includegraphics[width=60pt]{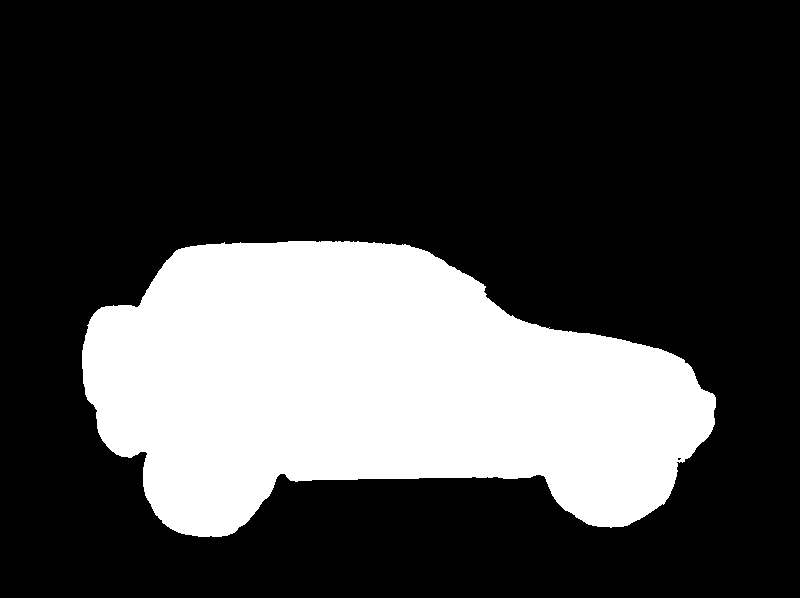} &
    \includegraphics[width=60pt]{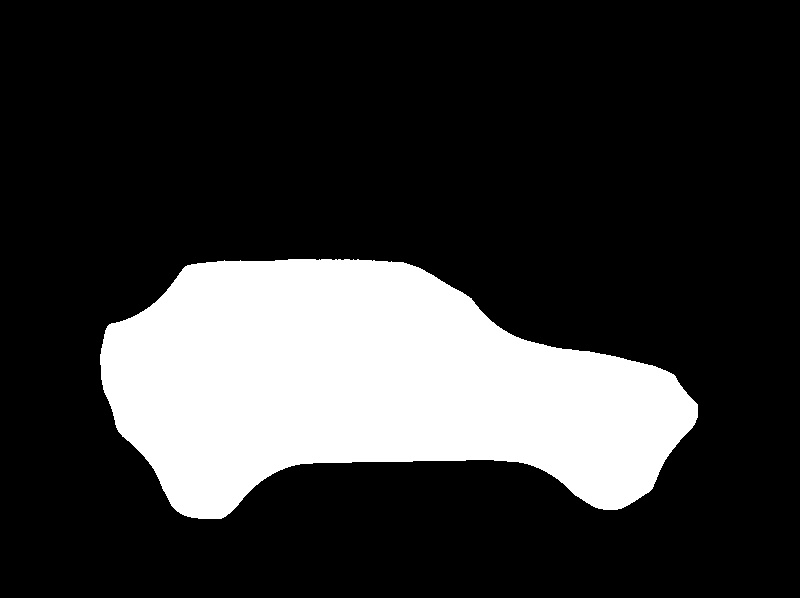} &
    \includegraphics[width=60pt]{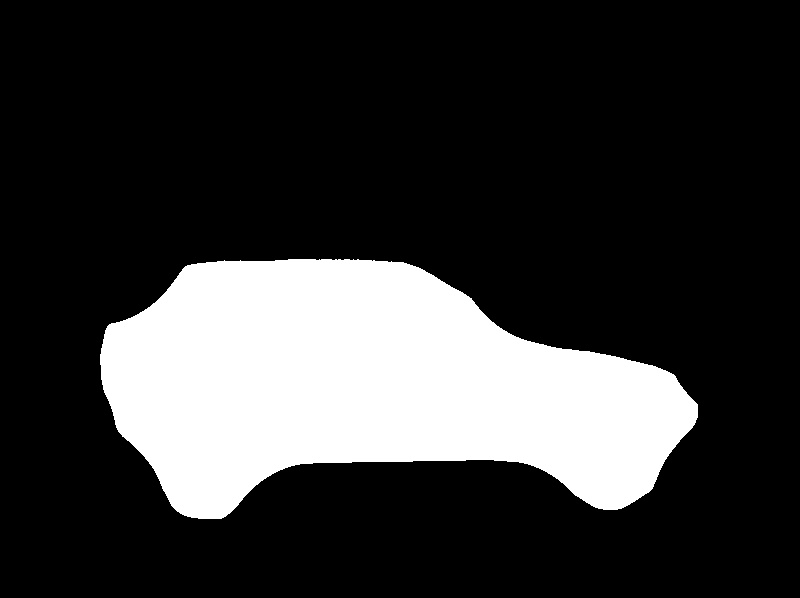} &
    \includegraphics[width=60pt]{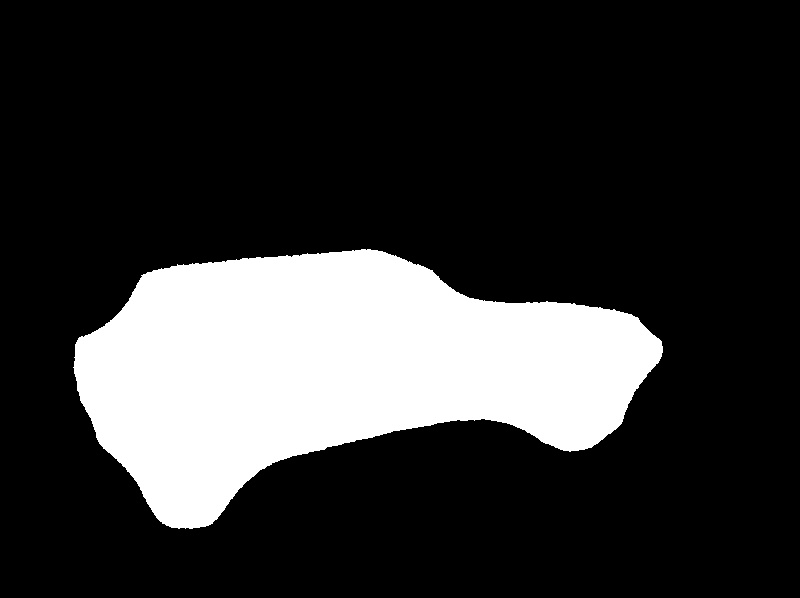} &
    \includegraphics[width=60pt]{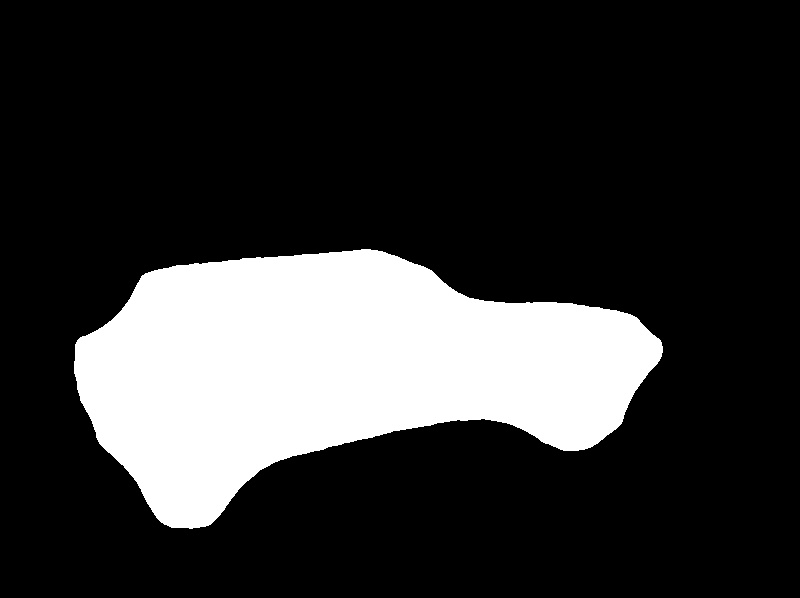} &
    \includegraphics[width=60pt]{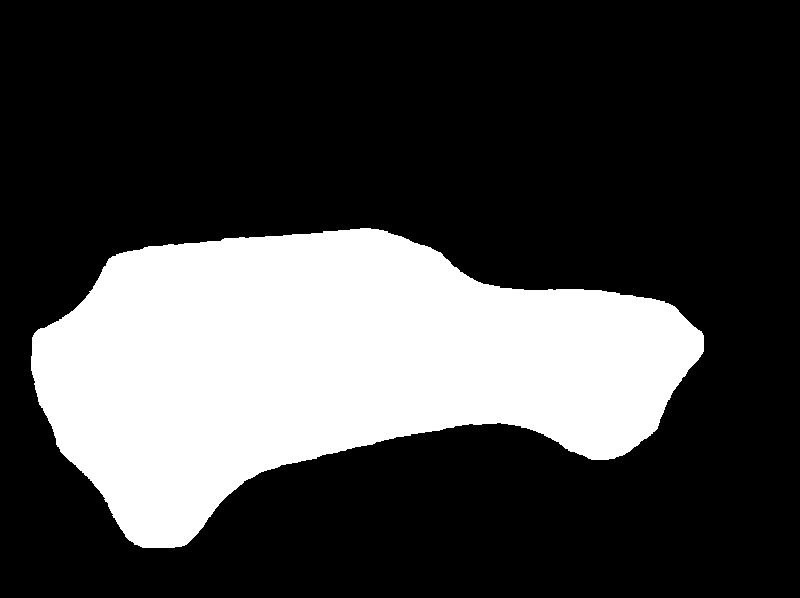} &
    \includegraphics[width=60pt]{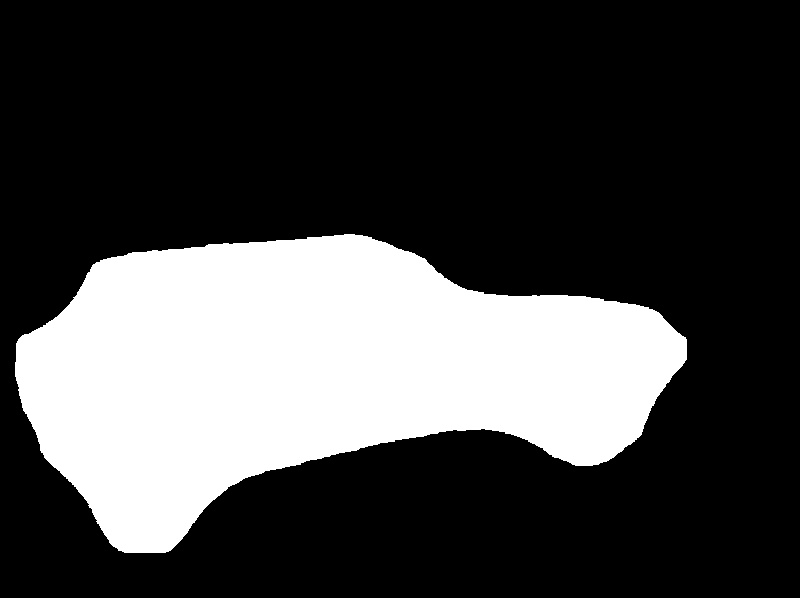} &
    \includegraphics[width=60pt]{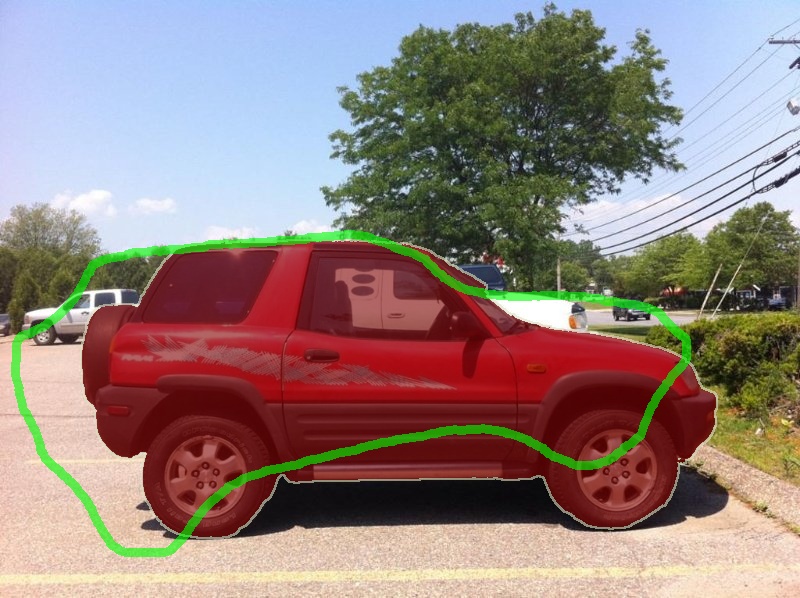} \\
    \includegraphics[width=60pt]{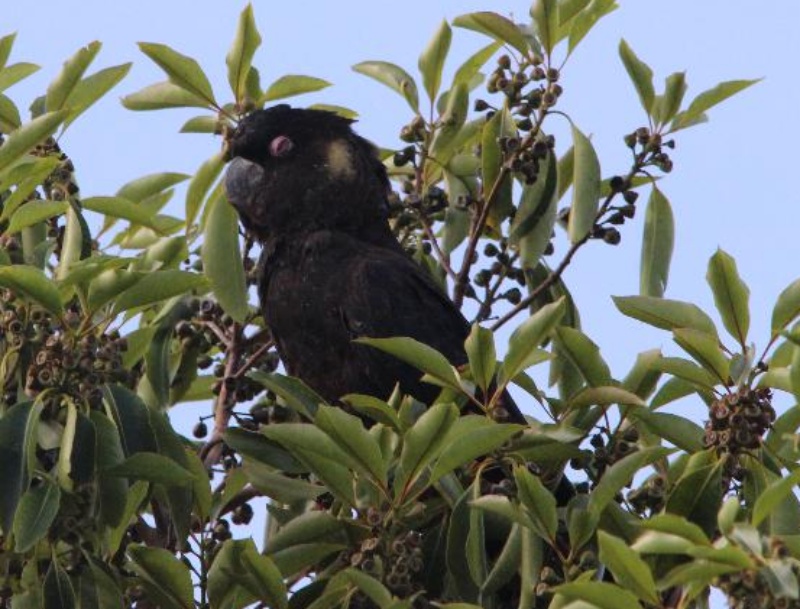} &
    \includegraphics[width=60pt]{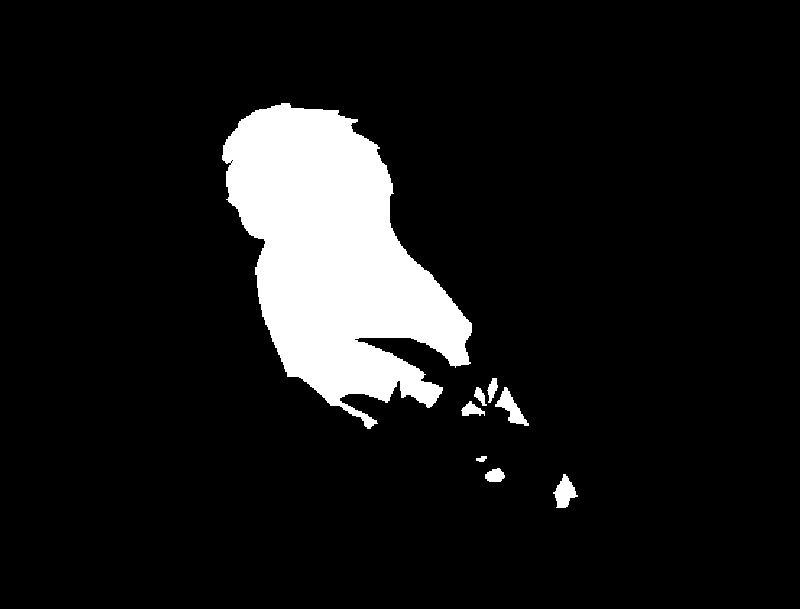} &
    \includegraphics[width=60pt]{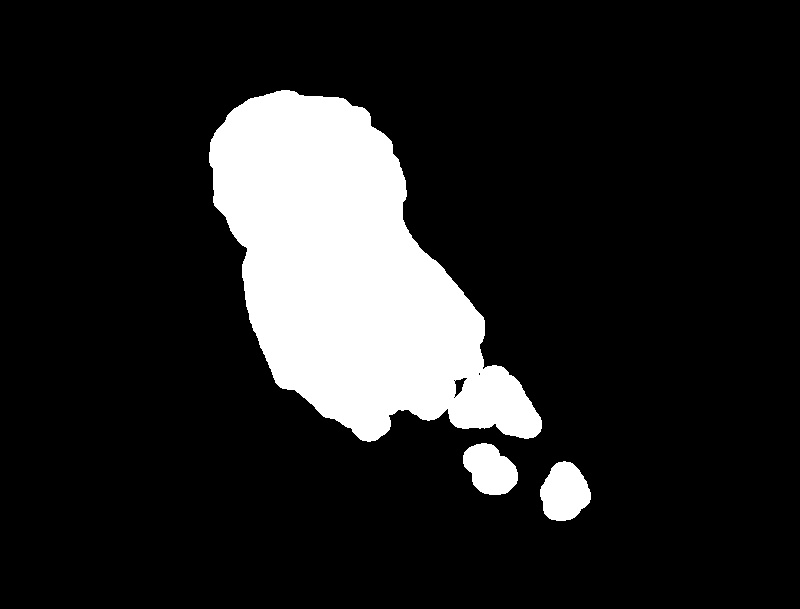} &
    \includegraphics[width=60pt]{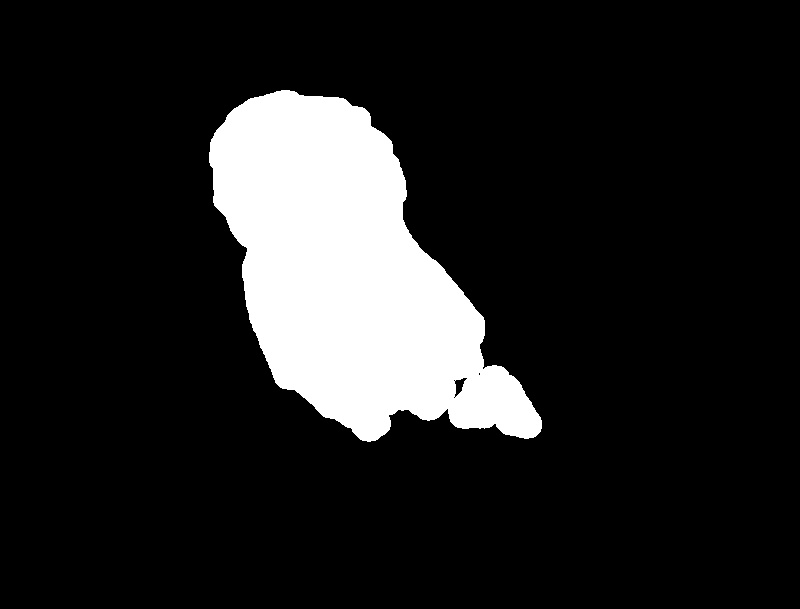} &
    \includegraphics[width=60pt]{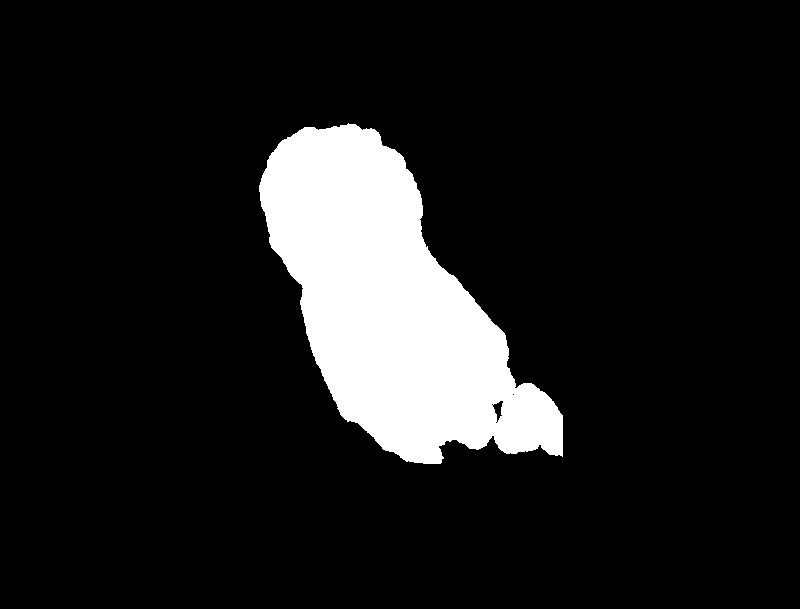} &
    \includegraphics[width=60pt]{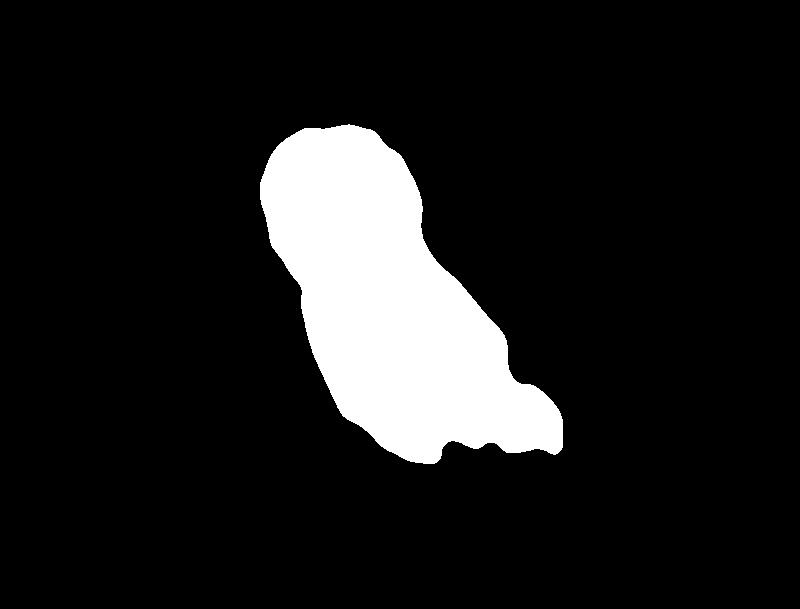} &
    \includegraphics[width=60pt]{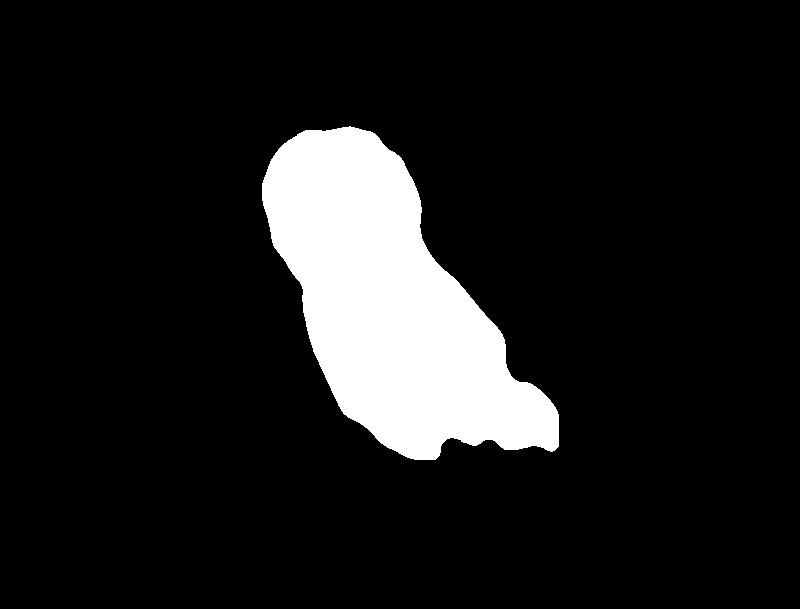} &
    \includegraphics[width=60pt]{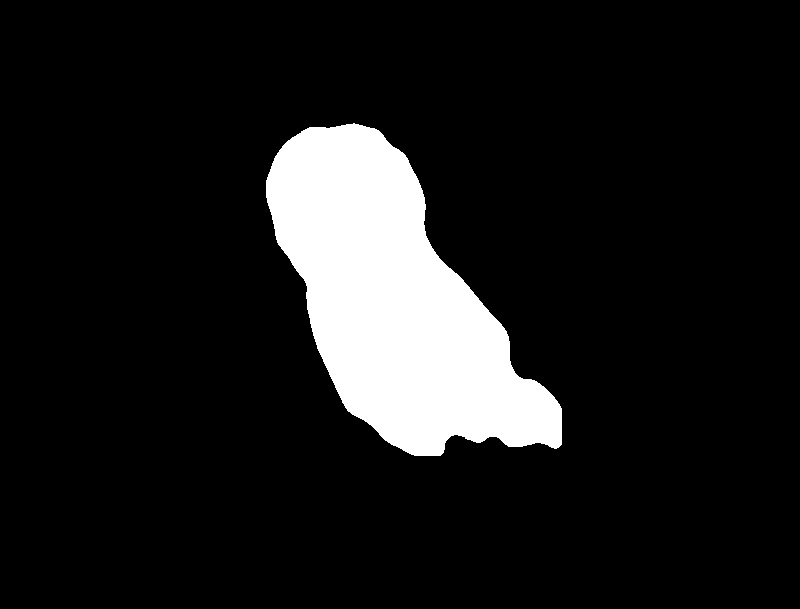} &
    \includegraphics[width=60pt]{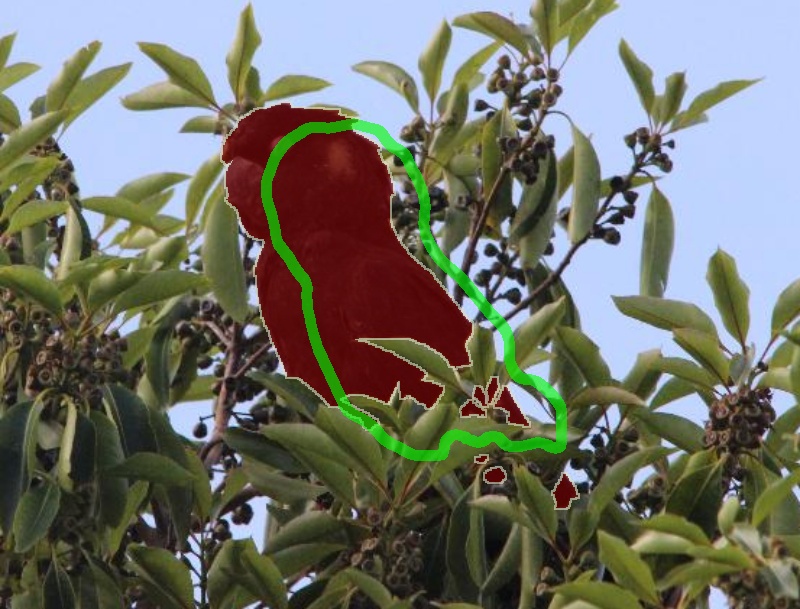} \\
    \includegraphics[width=60pt]{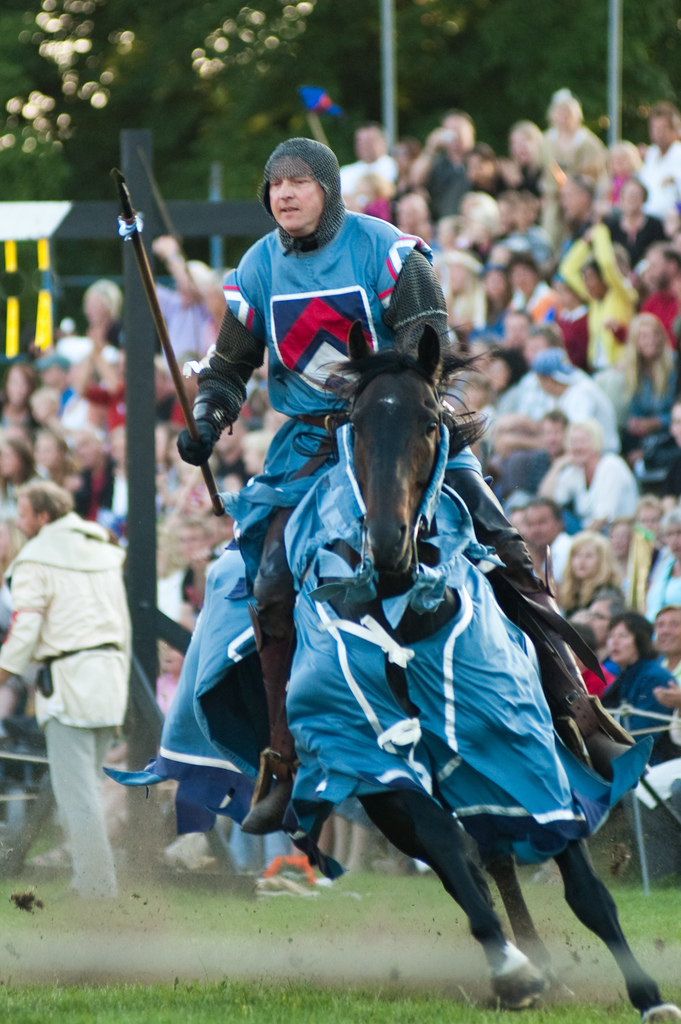} &
    \includegraphics[width=60pt]{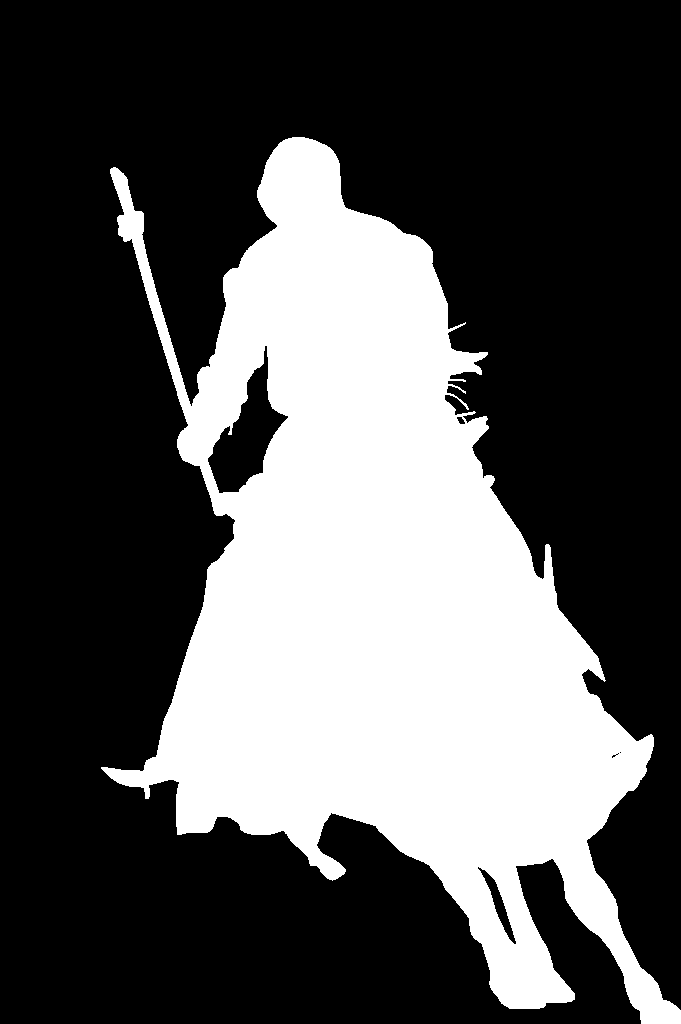} &
    \includegraphics[width=60pt]{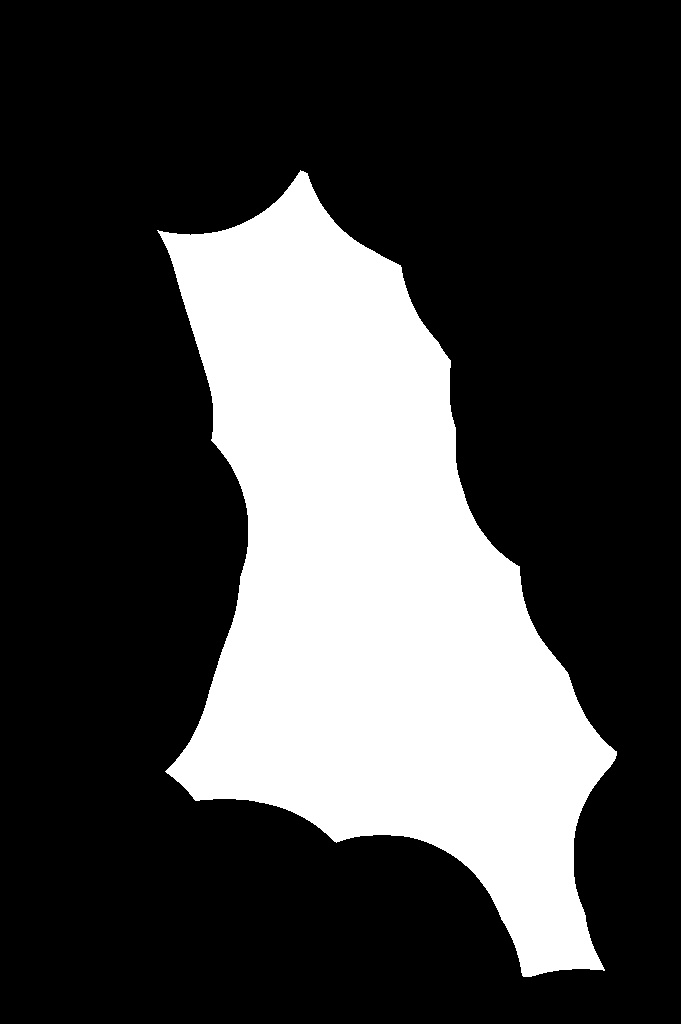} &
    \includegraphics[width=60pt]{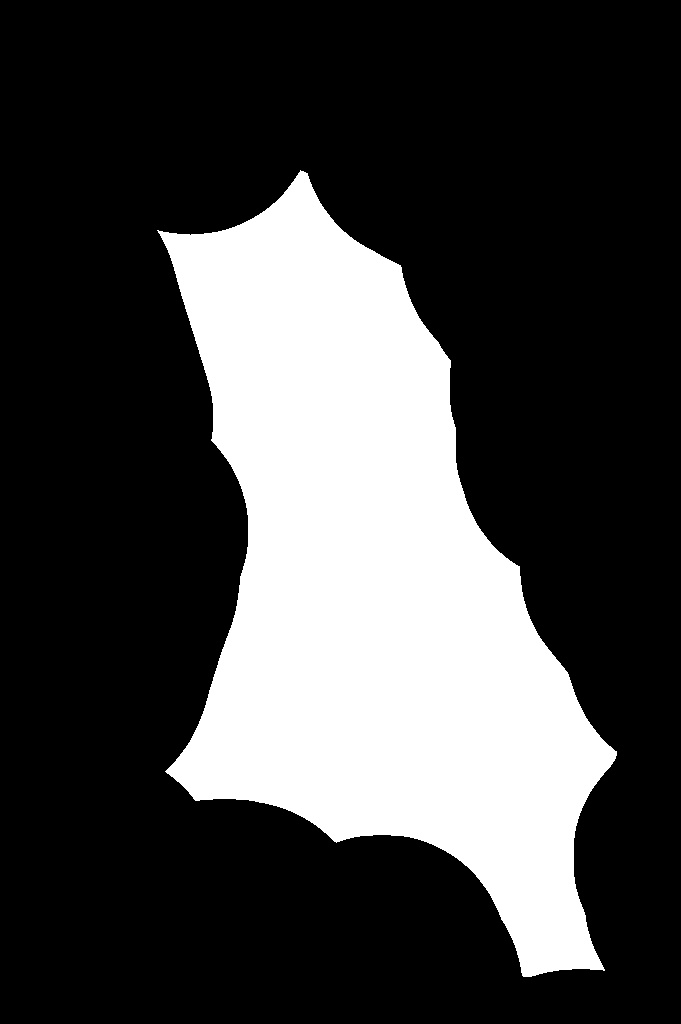} &
    \includegraphics[width=60pt]{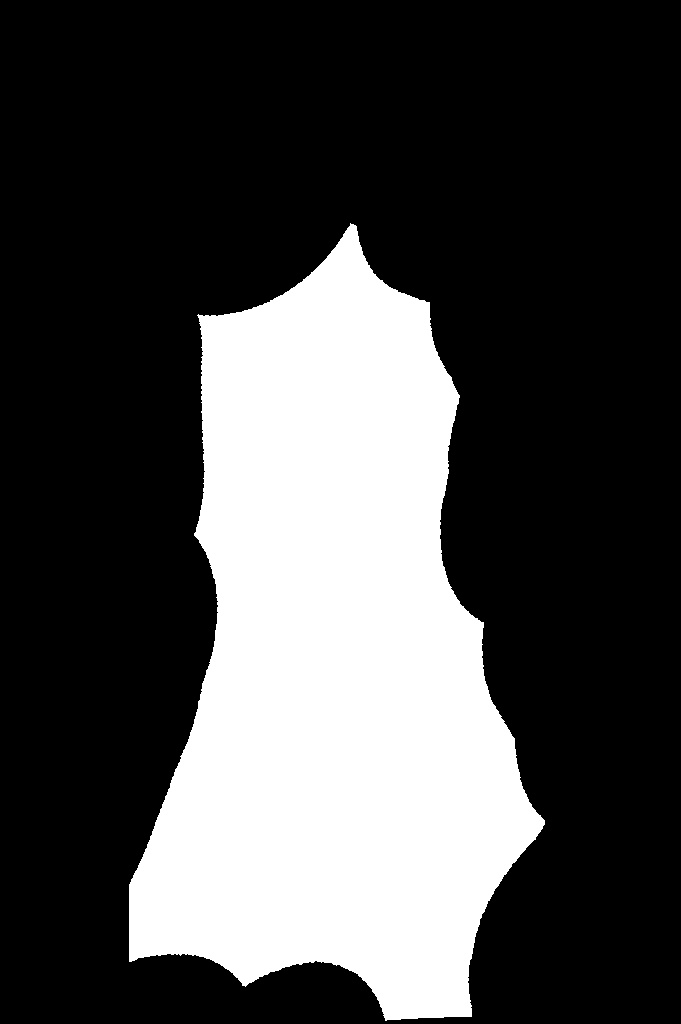} &
    \includegraphics[width=60pt]{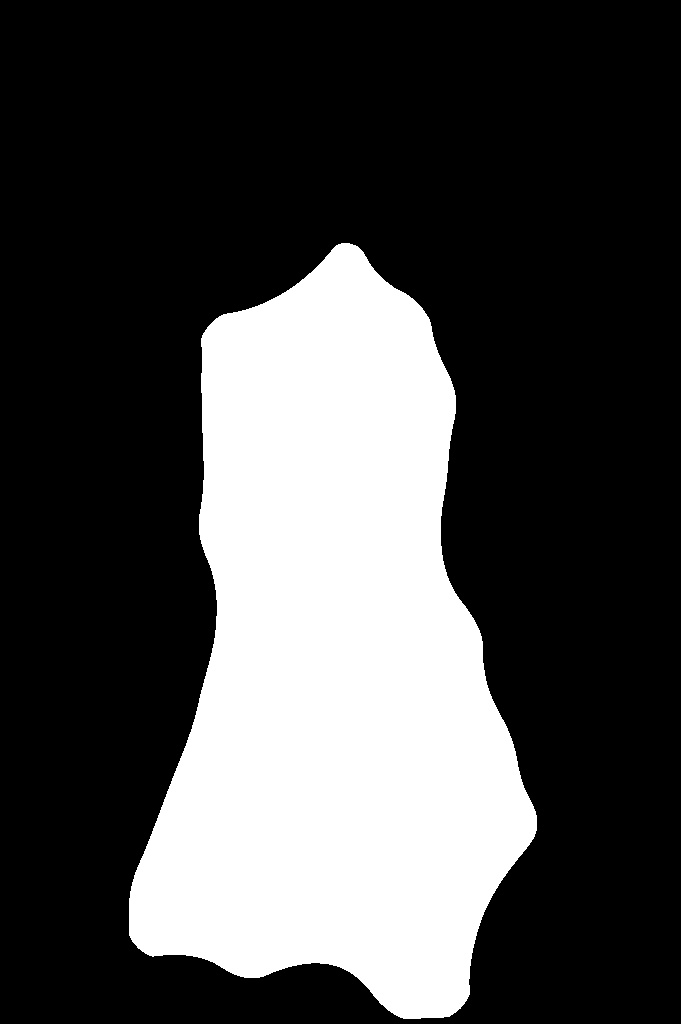} &
    \includegraphics[width=60pt]{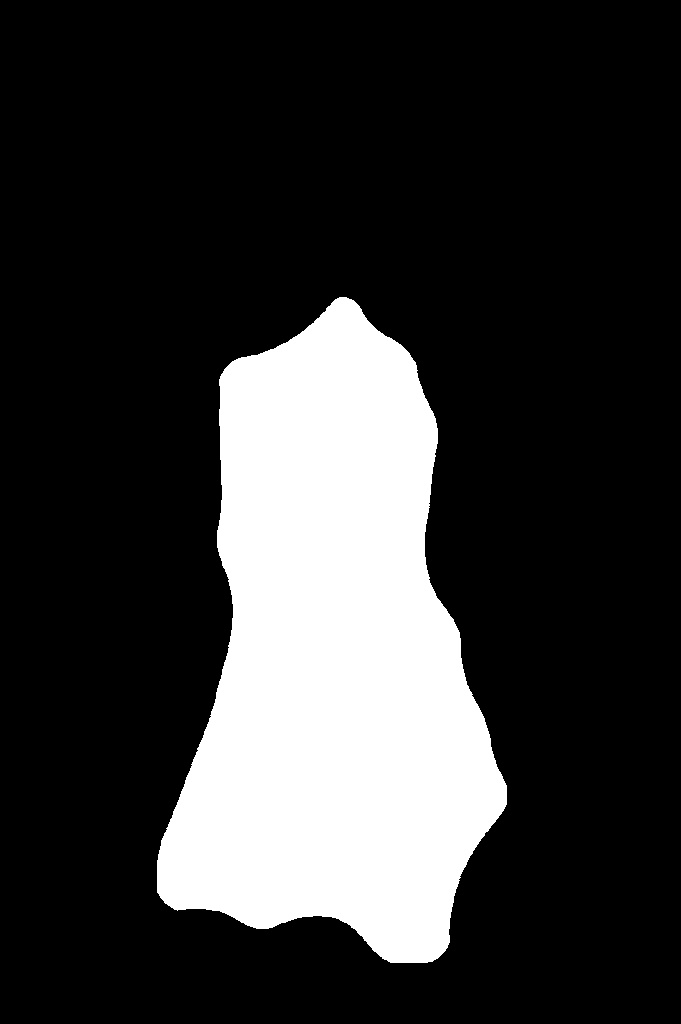} &
    \includegraphics[width=60pt]{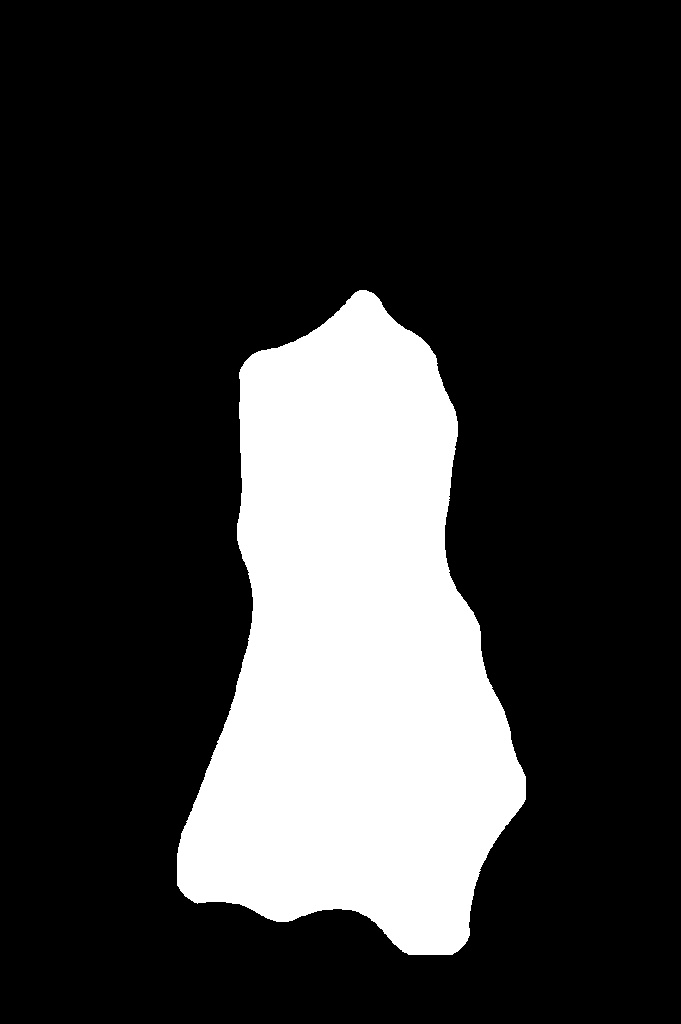} &
    \includegraphics[width=60pt]{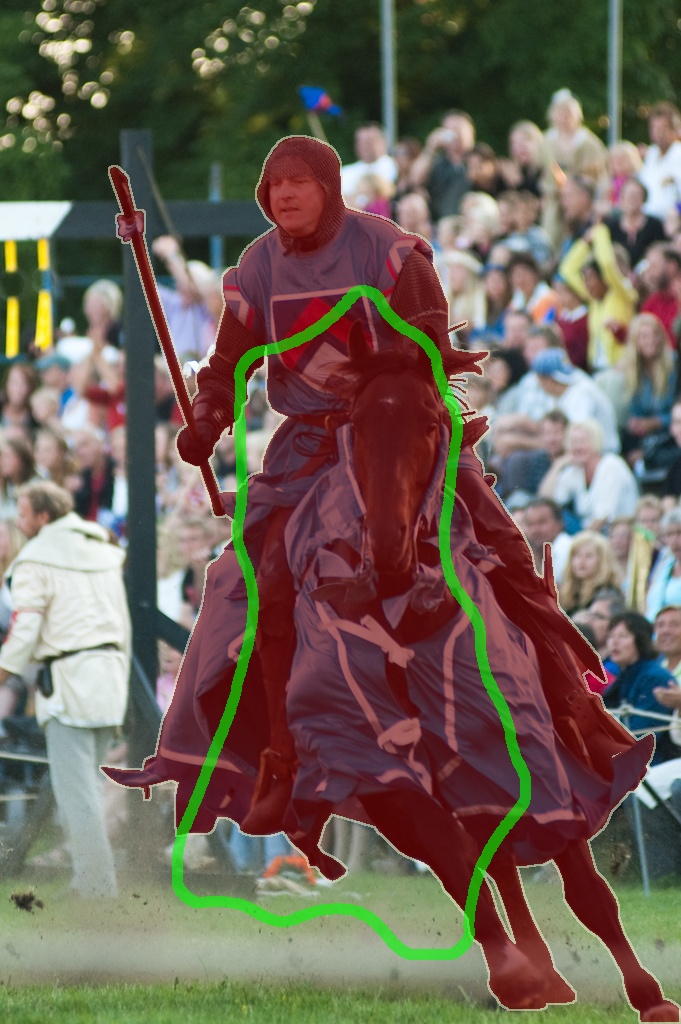} \\
\end{tabular}
}
\captionsetup{labelformat=empty}
\captionof{figure}{
  Figure 3: Step-by-step contour generation given a ground truth segmentation mask.
}
\label{fig:contours-generation-stepwise}
\end{center}
}
\end{table*}

\subsection{Contour Generation Module}

The contour generation module is designed to emulate real user behavior. As we are unaware of a user study of contouring patterns, we develop a generation procedure based on general assumptions. Our goal is that a network trained on generated samples performs well on real user inputs. 

\paragraph{Contour generation.}
A user is not expected to outline an object as accurate as possible: a real contour might cross object boundaries, do not cover some extruding object parts, or even lie within the object area instead of enclosing the object. Accordingly, to generate a contour from a ground truth instance mask, we need to distort it rather heavily. Overall, we generate contours via the following algorithm:

\begin{enumerate}
    \item First, we fill all holes in the mask. 

    \item After that, we randomly select either dilation or erosion. For a chosen morphological transform, we randomly sample a kernel size depending on the image size. Hence, the transformed mask does not stretch or shrink too much, yet close-by objects might merge, so that the contour encloses a group of objects.

    \item A contour should not outline distant parts of an object or even different objects, so we do not consider disconnected areas of the transformed mask. So, we search for connected components and select the largest one. 

    \item Then, we distort the mask via an elastic transform with random parameters depending on the object size. This might divide the mask into several disconnected parts, so we select the largest connected component yet again.

    \item We smooth the mask via GaussianBlur with a random kernel size chosen according to the object size.

    \item Next, we apply random scaling. We assume objects of a simple shape might be outlined rather coarsely, while complex shapes require a more thoughtful approach. Accordingly, we define a ratio $r$ reflecting the “complexity” of the object shape: it is calculated as the area of the current mask divided by the area of its convex hull. If $r < 0.6$, we assume an object has a complex, non-convex shape. In this case, we cannot apply severe augmentations to the mask, since the distorted mask would match the object badly. If $r \ge 0.6$, an object seems to be “almost convex”, so intense augmentations would not affect its shape so dramatically. Accordingly, we randomly sample a scaling factor within a narrow range for complex, “non-convex” objects, and from a wider range for less complex, “almost convex” objects.

    \item Finally, we select a shift based on the object size. Particularly, we consider bounding boxes enclosing the transformed and the ground truth masks, and compute $d_x$ as a minimum distance along $x$-axis between vertical sides of the ground truth and transformed boxes. An integer shift along the $x$-axis is sampled from $[-2d_x, 2d_x]$ for “almost convex” or $[-d_x, d_x]$ for “non-convex” objects, respectively. A shift along the $y$-axis is selected similarly. The resulting mask defines a filled contour. 

\end{enumerate}

To clarify our generation procedure, we visualize the intermediate results in Fig.~\ref{fig:contours-generation-stepwise}. More details, including hyperparameter values, are provided in Supplementary.

Fig.~\ref{fig:heatmaps} depicts multiple contours generated for a single mask. Generated contours may vary in size and shape significantly due to the randomized generation procedure.

\begin{table}[h!]
\setlength{\tabcolsep}{1pt}{
\begin{center}
\resizebox{0.95\linewidth}{!}{
\begin{tabular}{cccc}
    \small
    GT mask & \multicolumn{2}{c}{Random contours} & Contour heatmap \\
    \includegraphics[height=90pt]{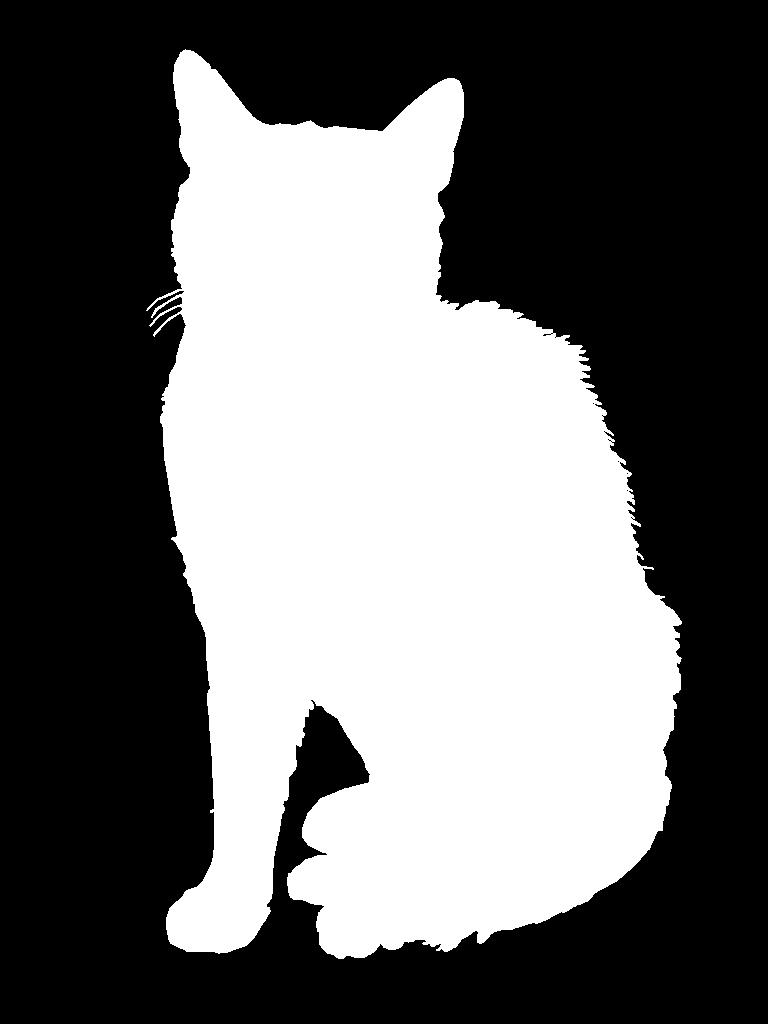} & 
    \includegraphics[height=90pt]{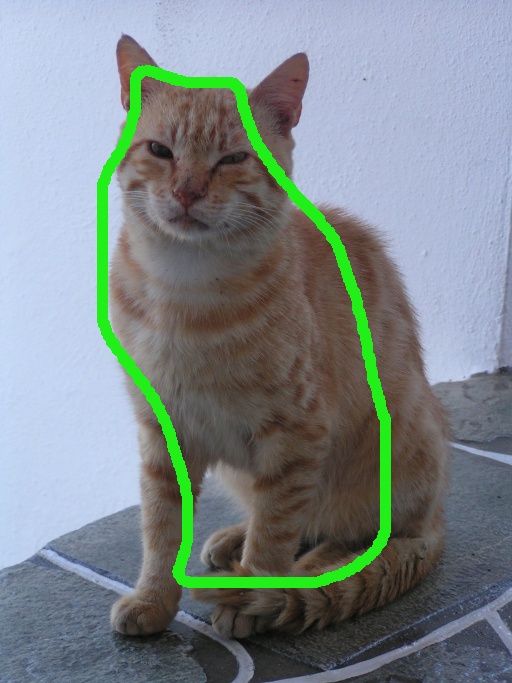} &
    \includegraphics[height=90pt]{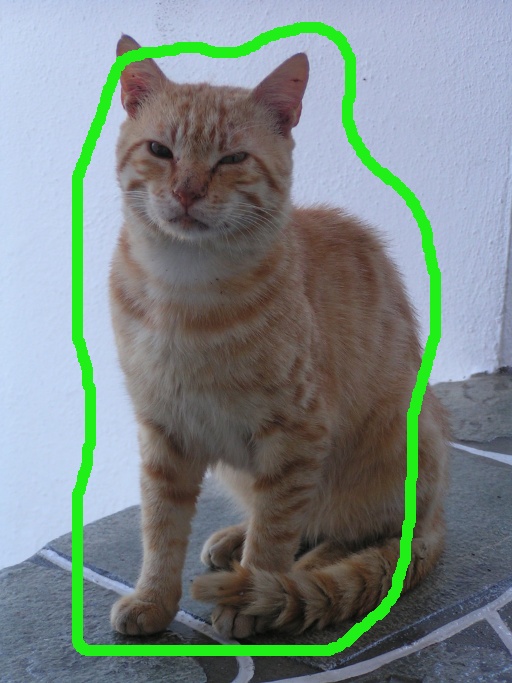} &
    \includegraphics[height=90pt]{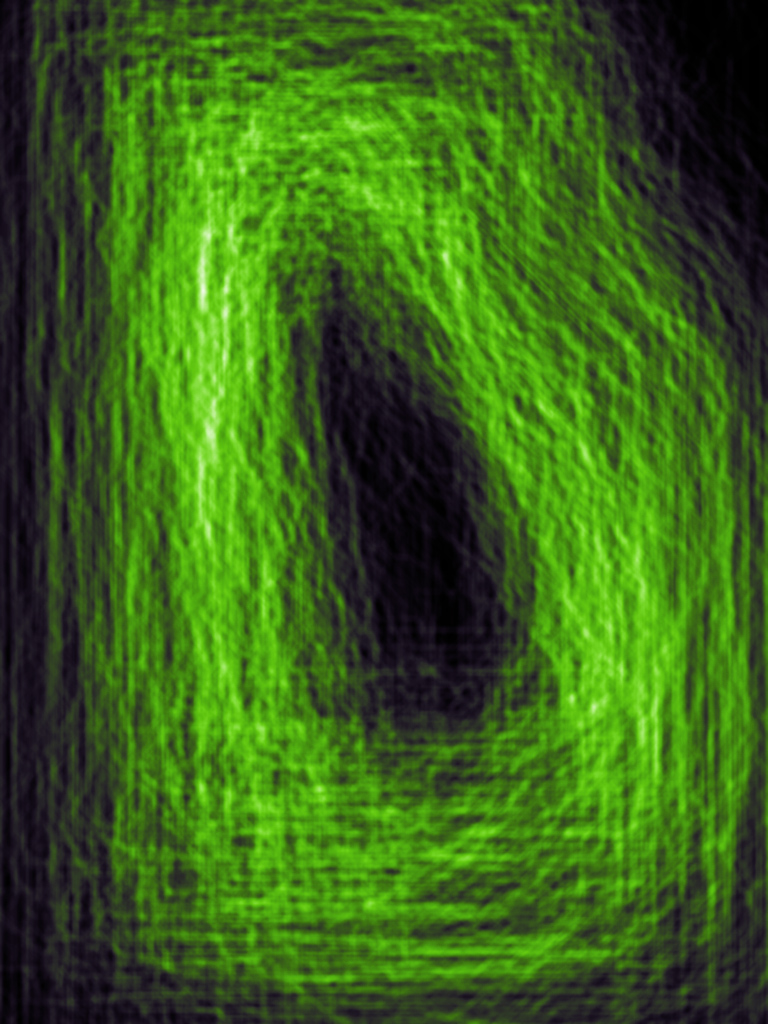}
\end{tabular}
}
\captionsetup{labelformat=empty}
\captionof{figure}{
  Figure 4: To emulate real user input, we aim to generate as diverse contours as possible. The only requirement is that they should adequately represent the desired object and allow unambiguously identifying it. A heatmap with 1000 random contours illustrates the range of variation. Here, we visualize contours as lines for clarity.
}
\label{fig:heatmaps}
\end{center}
}
\end{table}

\paragraph{Contours encoding.}

We formulate contours encoding guided by clicks encoding in click-based approaches. According to the study on clicks encoding~\cite{benenson2019large}, the best way is to encode clicks as binary masks with click positions marked with disks of a fixed radius (as in RITM~\cite{sofiiuk2021reviving}). So we also represent contours as binary masks, considering two ways of encoding contours: “filled” contour masks with ones inside and zeros outside the contour, and contours drawn as lines. Our ablation study of contours encoding (Tab.~\ref{tab:ablation-encoding}) shows that “filled” contours provide higher segmentation quality than lines. We attribute this to filled masks providing explicit information about whether each pixel lies within a contour. Due to a limited receptive field, convolutional neural networks might not derive this information as effectively from contours drawn as lines.

Each user input is encoded as two binary masks: one for a positive contour and another for a negative one (one empty and one filled, depending on whether a positive or negative contour is drawn). We also leverage the history of user inputs contained in the mask predicted at the previous interaction. Prior to the first input, we pass an all-zeros mask. Overall, the network input is two binary maps stacked with the previous mask channel-wise, as in RITM~\cite{sofiiuk2021reviving}.

\subsection{Backbone}

Following RITM~\cite{sofiiuk2021reviving}, we use HRNet18 with an OCR module~\cite{WangSCJDZLMTWLX19,YuanCW19} as a backbone. We also examine other HRNet backbones: a lightweight HRNet18s and more powerful HRNet32 and HRNet48 models. The results evident that the network complexity has a minor effect on the segmentation quality (Tab.~\ref{tab:ablation-backbone}).

\subsection{Interactive Branch}

We modify a segmentation network by adding an interactive branch that processes an additional user input. This is implemented via Conv1S, a network modification proposed in RITM~\cite{sofiiuk2021reviving}. Specifically, we pass the interaction through the interactive branch made up of Conv+LeakyReLU+Conv layers. Then, we sum up the result with the output of the first convolutional backbone layer. Yet, we observe that the interactive branch output might confuse the network at the beginning of the training. To avoid this, we extend the interactive branch with a scaling layer, that multiplies the output by a learnable coefficient just before summation. Through scaling, we can balance the relative importance of image features and user inputs in a fully data-driven way.

\section{Experiments}

\subsection{Standard Benchmarks}

Following RITM~\cite{sofiiuk2021reviving}, we train our models on the standard segmentation datasets. Specifically, we use Semantic Boundaries Dataset, or SBD~\cite{hariharan2011semantic}, and the combination of LVIS~\cite{gupta2019lvis} and COCO~\cite{lin2014microsoft} for training. SBD~\cite{hariharan2011semantic} consists of 8498 training samples. COCO and LVIS share the same set of 118k training images; COCO contains a total of 1.2M instance masks of common objects, while LVIS~\cite{gupta2019lvis} is annotated with 1.3M instance masks with long-tail object class distribution. Respectively, the combination of COCO and LVIS contains small yet diverse set of classes from LVIS and general and large set of classes from COCO. In an ablation study of training data, we use the test+validation split of OpenImages~\cite{kuznetsova2020open} (about $100$k samples); we do not consider the train split since it is annotated quite inaccurately. 

We evaluate our method on standard IS benchmarks: GrabCut~\cite{rother2004grabcut} (50 samples), the test subset of Berkeley~\cite{martin2001database,mcguinness2010comparative} (100 samples), a set of 345 randomly sampled frames from DAVIS~\cite{perazzi2016benchmark,jang2019interactive}, and the test subset of SBD (539 samples). Originally, these benchmarks do not contain contour annotations, so we manually label them with contours by ourselves. 

\subsection{Proposed Datasets}

We present the \textbf{\dset{}} dataset with \dsetsize{} images depicting common objects in their natural context. Besides, we manually select \dsetgroupsize{} samples containing objects groups to create an especially challenging \textbf{\dsetgroup{}}. The examples of images along with instance segmentation masks and user-defined contours are shown in Fig.~\ref{fig:dataset-examples} and Fig.~\ref{fig:groupdataset-examples}.

\paragraph{Source of data.} The images of \dset{} are taken from the train subset of OpenImages V6. We selected \dsetsize{} diverse images depicting common objects in various scenarios. The only restriction is imposed on image resolution: we consider only images with a shorter side of between $400\text{px}$ and $1600\text{px}$, and of an aspect ratio between 1:5 and 5:1.

\begin{table}[h!]
\setlength{\tabcolsep}{1pt}{
\begin{center}
\resizebox{0.95\linewidth}{!}{
\begin{tabular}{ccc}
    \includegraphics[height=80pt]{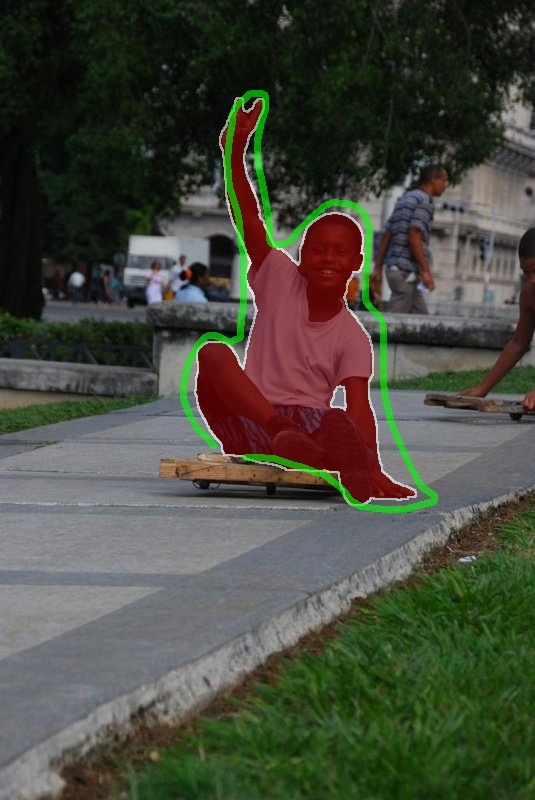} &
    \includegraphics[height=80pt]{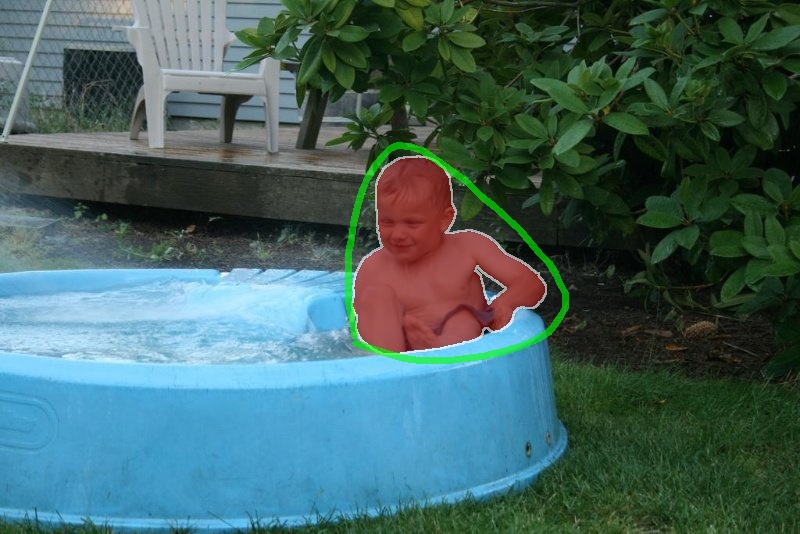} & 
    \includegraphics[height=80pt]{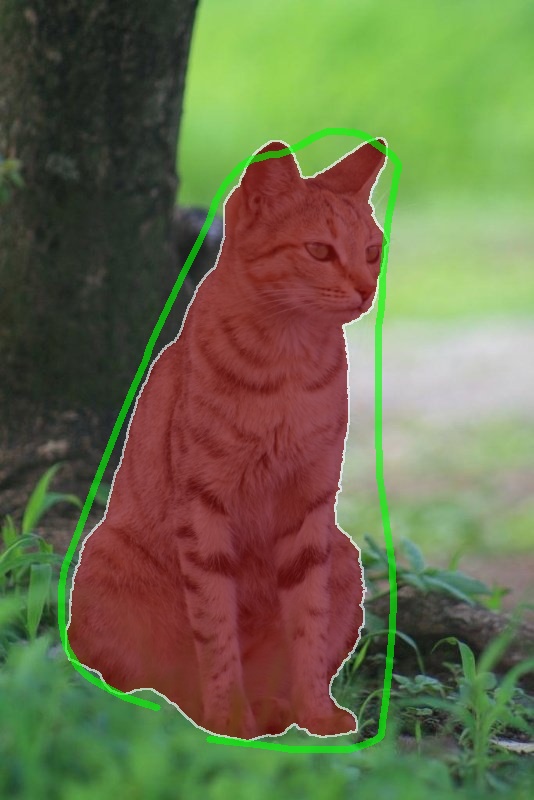}
\end{tabular}
}
\captionsetup{labelformat=empty}
\captionof{figure}{
  Figure 5: Examples from \dset. User-defined contours are green, instance segmentation masks are red. Contours might be loose and non-closed.
}
\label{fig:dataset-examples}
\end{center}
}
\end{table}

\paragraph{Instances labeling.} 

We decomposed the labeling task into two subtasks. The first subtask implies creating instance segmentation masks for the given images. For the test subsets of the standard benchmarks, we use pre-defined instance segmentation masks already present in these datasets. The second subtask is to outline instances with contours.

\begin{table}[h!]
\setlength{\tabcolsep}{1pt}{
\begin{center}
\resizebox{0.95\linewidth}{!}{
\begin{tabular}{ccc}
    \includegraphics[height=110pt]{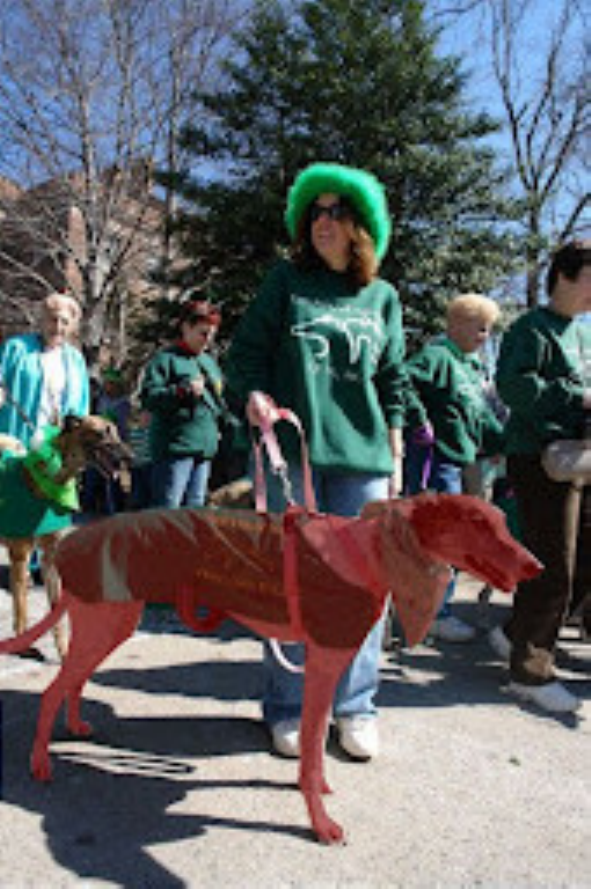} &
    \includegraphics[height=110pt]{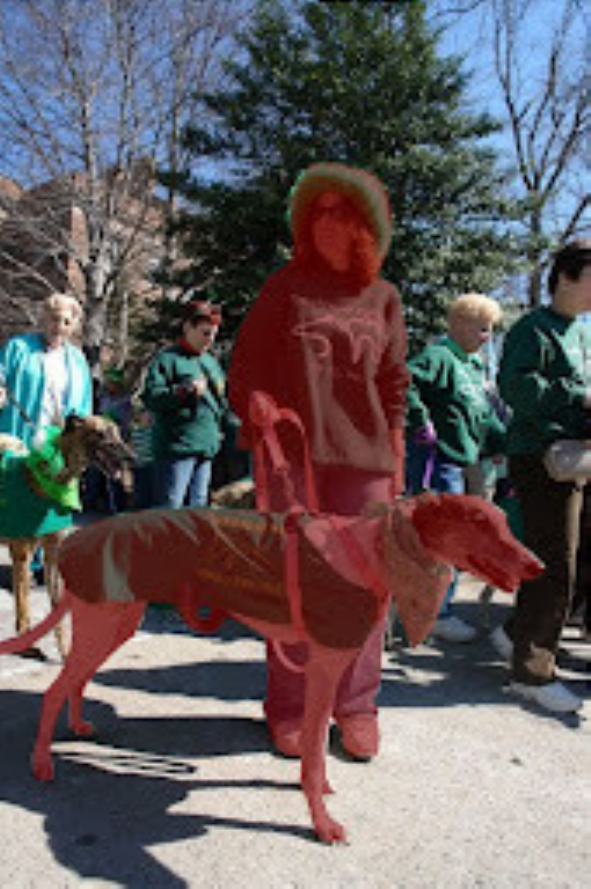} &
    \includegraphics[height=110pt]{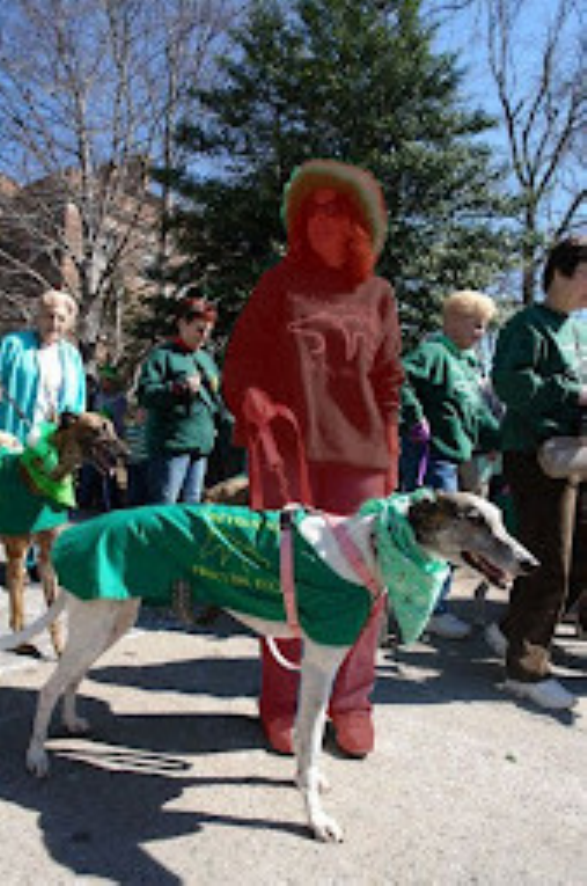} \\
\end{tabular}
}
\captionsetup{labelformat=empty}
\captionof{figure}{
  Figure 6: The instances (marked with red) segmented by different users in the same image might also be different.
}
\label{fig:unambiguous-definition}
\end{center}
}
\end{table}

First, we label images with instance masks. Since either a single object or a group of close-by objects can be a subject of interest, we request our annotators to label 50\% of instance masks as groups and 50\% as individual objects. At least one object per image should be annotated. To make the annotation process more alike with real image editing scenarios, we do not explicitly formulate what is a desired object. Instead, we ask annotators to label any objects that stand out (Fig.~\ref{fig:unambiguous-definition}). We also do not restrict the object size or the location in an image. The annotation guide and examples of annotations can be found in Supplementary.

\paragraph{Contours labeling.}

We asked our annotators to outline each segmented instance with a \textit{contour}: a line loosely following object boundaries. There should be no intermediate breaks in a contour; however, its start may not coincide with its end: it this case, we close the contour by connecting the first and the last points with a line. We aim to emulate real user interactions, so we requested for the contours that are not as precise as possible, but drawn in a natural relaxed manner. Nevertheless, the correspondence between instances and contours should be clear and unambiguous. Negative contours might be used only when necessary (Fig.~\ref{fig:negative-contours}).

\begin{table}[h!]
\setlength{\tabcolsep}{1pt}{
\begin{center}
\resizebox{0.95\linewidth}{!}{
\begin{tabular}{ccc}
    \small
    \includegraphics[height=80pt]{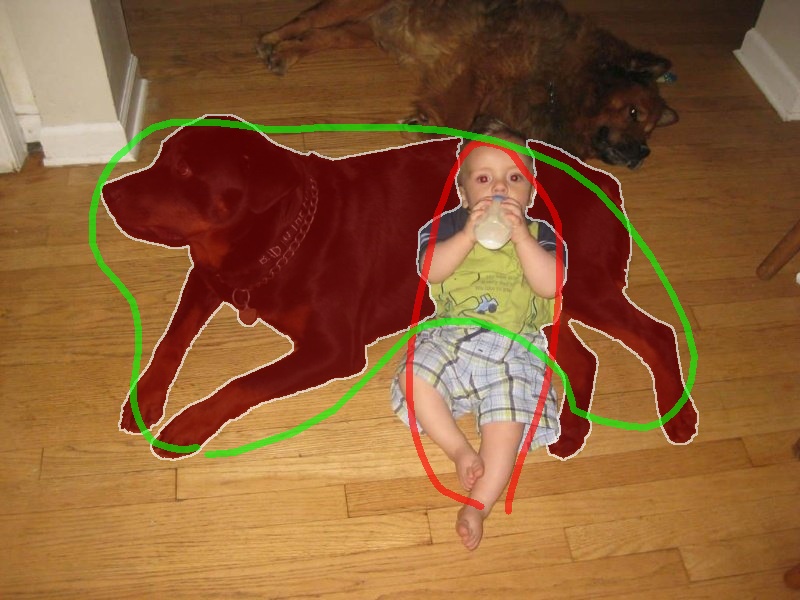} &
    \includegraphics[height=80pt]{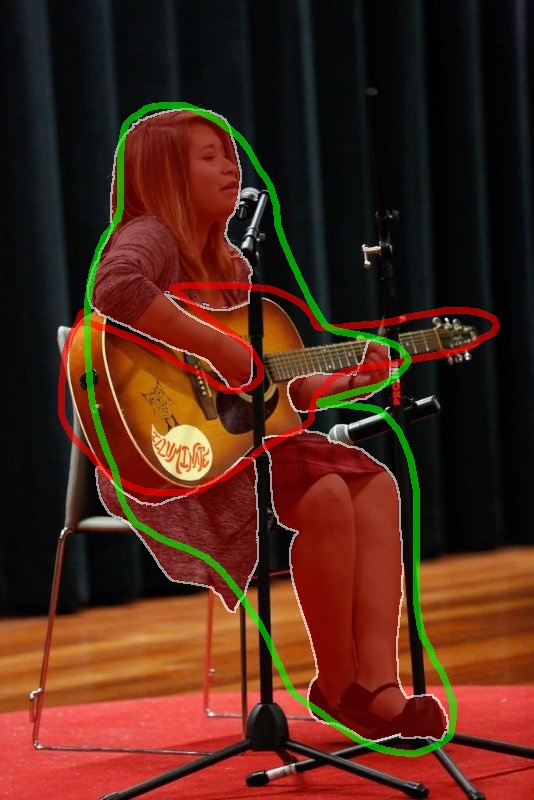} & \includegraphics[height=80pt]{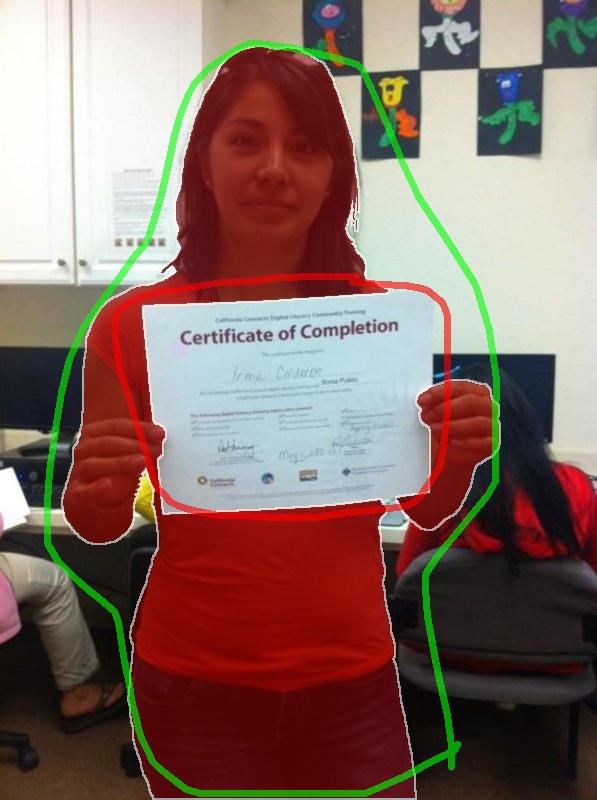} 
\end{tabular}
}
\captionsetup{labelformat=empty}
\captionof{figure}{
  Figure 7: If an overlapped object cannot be selected with a single contour, negative contours are allowed. Positive contours are green, negative are red.
}
\label{fig:negative-contours}
\end{center}
}
\end{table}

\paragraph{\dsetgroup.} We hand-picked \dsetgroupsize{} images depicting groups of objects from \dset{} to create a small yet extremely complex \dsetgroup{} (Fig.~\ref{fig:groupdataset-examples}).

\begin{table}[h!]
\setlength{\tabcolsep}{1pt}{
\begin{center}
\resizebox{0.95\linewidth}{!}{
\begin{tabular}{ccc}
    \includegraphics[height=54pt]{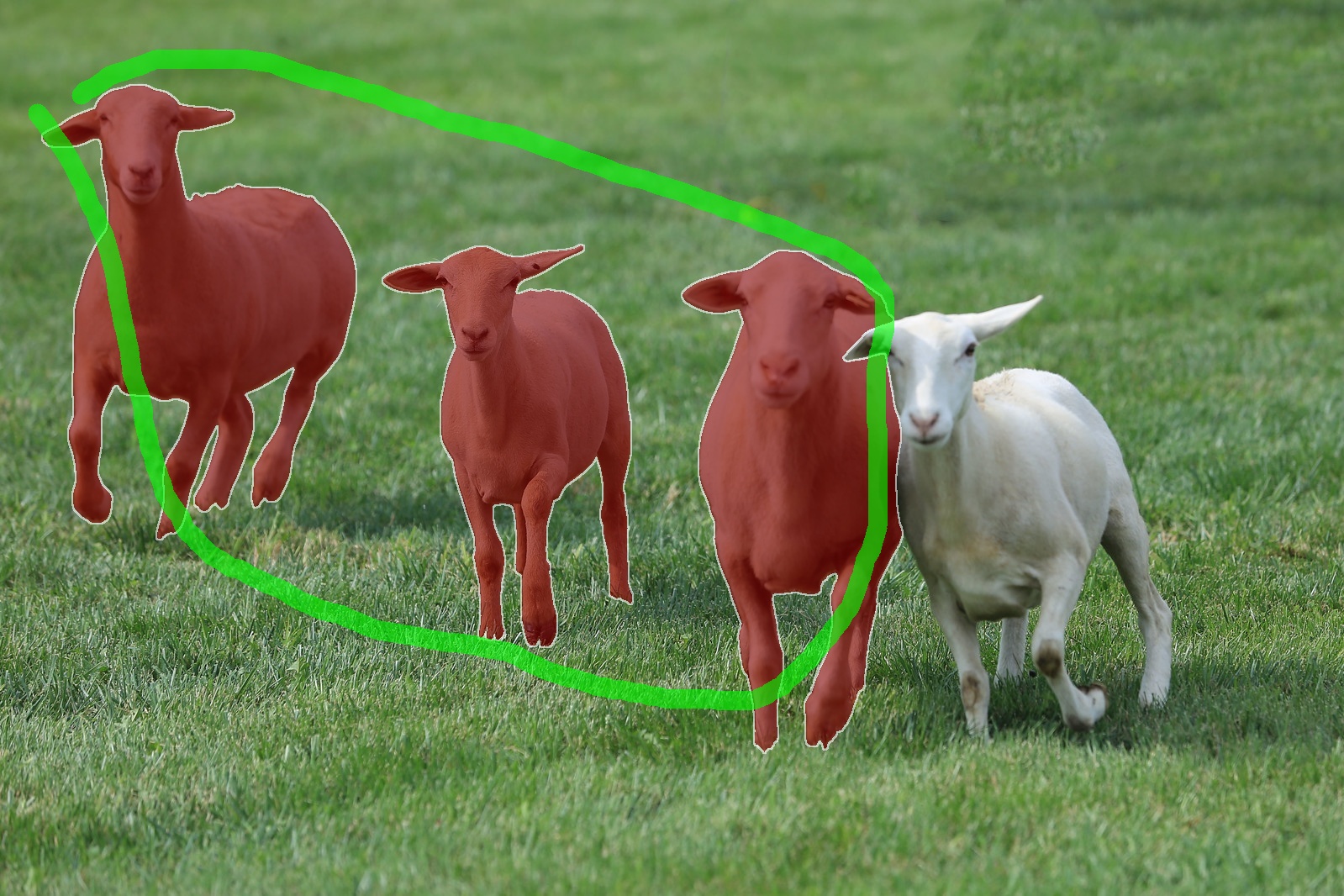} &
    \includegraphics[height=54pt]{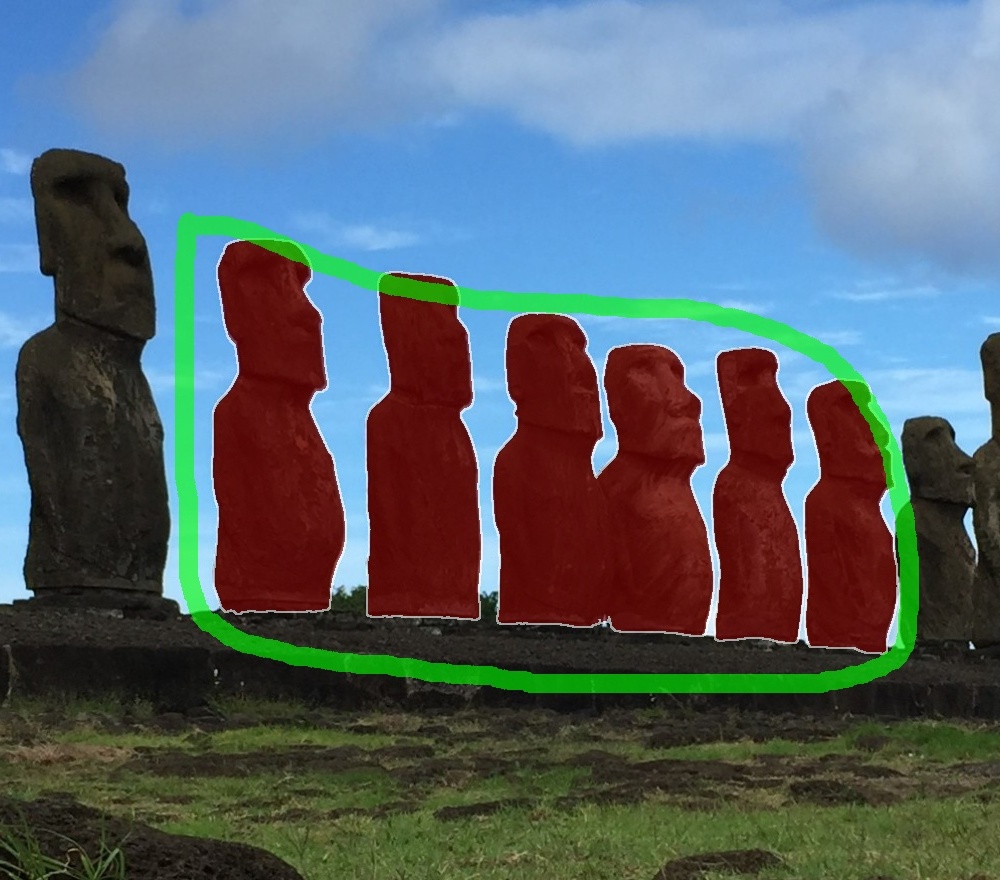} &
    \includegraphics[height=54pt]{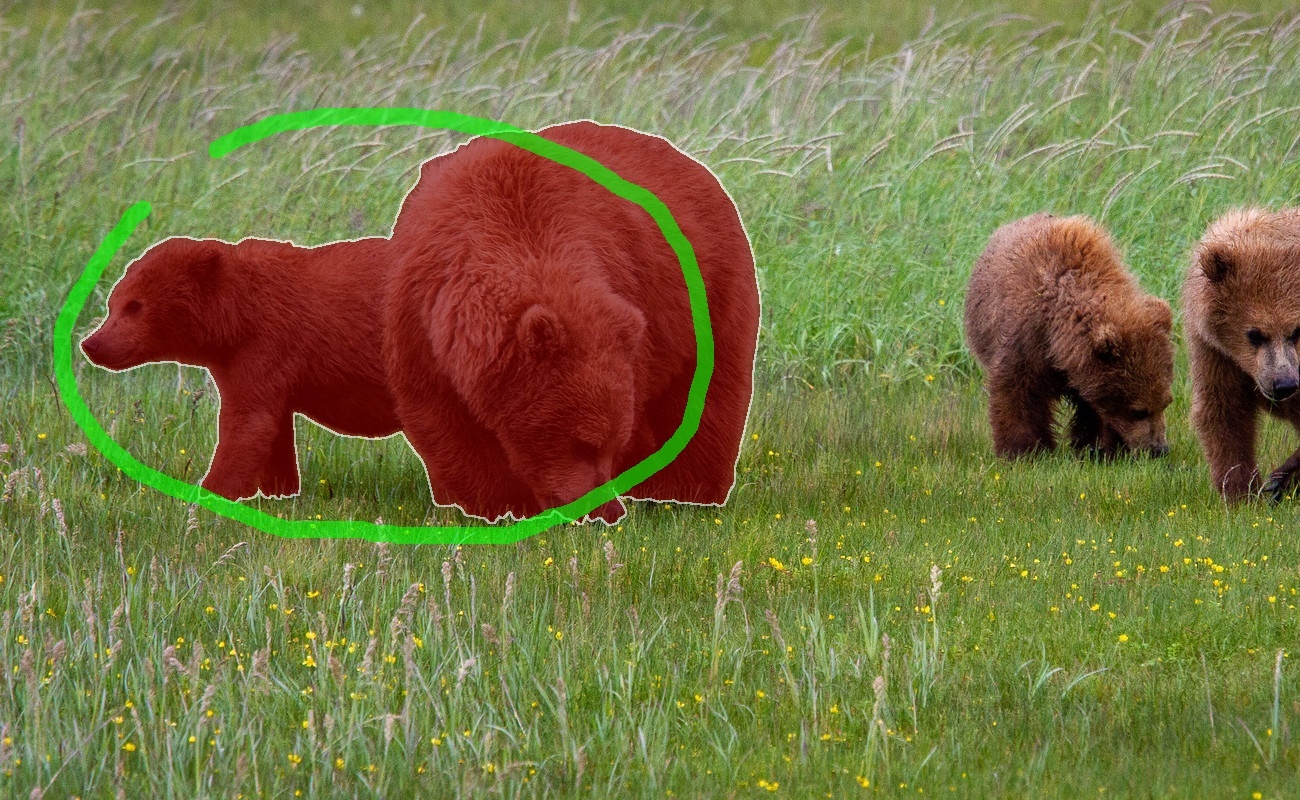} \\   
\end{tabular}
}
\captionsetup{labelformat=empty}
\captionof{figure}{
  Figure 8: Challenging images containing object groups from \dsetgroup.
}
\label{fig:groupdataset-examples}
\end{center}
}
\end{table}

\subsection{Evaluation}

\paragraph{Click-based evaluation.} Click-based IS methods are typically evaluated with NoC@k -- the number of clicks made to achieve a predefined IoU=k~\cite{hao2021edgeflow,jang2019interactive,Man+18,sofiiuk2020f,sofiiuk2021reviving}. For contours, the equivalent number of contours can be reported. However, this seems controversial as there exists a crucial difference between clicks and contours. A click is a pair of coordinates with a fixed and limited complexity, while a contour might be an arbitrarily long curve of an extremely complex shape. Since there is no conventional approach to measuring the curve complexity, we cannot formulate the relative complexity of clicks and contours and explicitly incorporate it into the evaluation metrics (e.g., in the form of scaling coefficients). Neither, we cannot treat clicks and contours equally.

\paragraph{Contour-based evaluation.} Since the proposed method is the first contour-based approach, there are no established evaluation protocols. Accordingly, we formulate an evaluation metric based on observations of human behavior. A user is not assumed to spend much time on image editing in mobile applications, so speed is a crucial factor of usability. Since the speed depends on the number of interactions, the fewer interactions, the better. Accordingly, it seems important to provide decent predictions even after the first contour. Since we manually annotate test datasets with contours, we can calculate IoU achieved with a single contour and use it as the main metric for assessing the contour-based IS. Or, we can apply click-based models and find the number of clicks required to achieve the same accuracy as achieved with a single contour. This way, contour-based methods can be compared with click-based methods non-directly.

\subsection{Implementation Details}

\paragraph{Training.} We train a binary segmentation model using a BCE loss. Input images are resized to $320\text{px} \times 480\text{px}$. During training, we randomly crop and rescale images, use horizontal flip, and apply random jittering of brightness, contrast, and RGB values. With an equal probability, we choose an object of interest with a positive contour or erase the unwanted object with a negative contour (passing the ground truth object mask as the previous mask, and treating the generated contour as a negative one).

The models are trained for 140 epochs using Adam~\cite{Kingma2014Adam} with $\beta_1 = 0.9$, $\beta_2=0.999$ and $\varepsilon = 10^{-8}$. The learning rate is initialized with $5 \cdot 10^{-4}$ and reduced by a factor of 10 at epochs 119 and 133.

In a study of training data, we fine-tune our models for 10 epochs. We use stochastic weight averaging~\cite{izmailov2018averaging}, aggregating the weights at every second epoch starting from the fourth epoch. During fine-tuning, we set the learning rate to $1 \cdot 10^{-5}$ for the backbone and $1 \cdot 10^{-4}$ for the rest of the network, and reduce it by a factor of 10 at epochs 8 and 9. 

\paragraph{Evaluation.}

We follow RITM~\cite{sofiiuk2020f} for the evaluation, using Zoom-In and averaging predictions from the original and the horizontally flipped images. 
Unlike a single click, a single contour allows hypothesizing about the object size, so we can apply Zoom-In at the first interaction; this minor change tends to improve the results significantly.

More information about software and hardware used in our experiments can be found in Supplementary materials.

\subsection{Comparison with Previous Works}

We present quantitative results for GrabCut (Tab.~\ref{tab:comparison-grabcut}), Berkeley (Tab.~\ref{tab:comparison-berkeley}), and DAVIS (Tab.~\ref{tab:comparison-davis}). We report a mean IoU for from 1 to 5 clicks for the click-based models, and an IoU after the first interaction for our contour-based models. Apparently, a single contour provides the same accuracy as 5 clicks on GrabCut and 3 clicks on Berkeley and DAVIS, compared with the state-of-the-art RITM. 

\begin{table}[h!]
    \begin{center}
    \tabcolsep=0.1cm
    \resizebox{1\linewidth}{!}{
    \begin{tabular}{l|c|ccccc}
        \hline
        \multirow{2}{*}{Method} & Training & \multicolumn{5}{c}{IoU} \\
        & data & @1 & @2 & @3 & @4 & @5 \\
        \hline
        BRS~\cite{jang2019interactive} & SBD & 80.0$^*$ & 87.0$^*$ & 89.0$^*$ & 90.0$^*$ & 90.0$^*$ \\
        f-BRS~\cite{sofiiuk2020f} & SBD & 80.0$^*$ & 85.0$^*$ & 87.0$^*$ & 91.0$^*$ & 92.0$^*$ \\
        EgdeFlow~\cite{hao2021edgeflow} & LC & 85.0$^*$ & 92.0$^*$ & 94.0$^*$ & 95.5$^*$ & \underline{96.5$^*$} \\
        RITM~\cite{sofiiuk2021reviving} & LC & 87.46 & 91.76 & 95.39 & 96.28 & \underline{97.20} \\
        \hline
        Ours & SBD & 96.42 & & & &  \\
        Ours & LC & 96.32 & & & &  \\
        \hline
    \end{tabular}
    }
    \end{center}
    \caption{A quantitative comparison of IS methods on GrabCut. LC denotes LVIS+COCO.  “*” means an approximate metric value determined from the IoU plots from the original papers. The results better than ours are \underline{underlined}.}
\label{tab:comparison-grabcut}
\end{table}

\begin{table}[h!]
    \begin{center}
    \tabcolsep=0.1cm
    \resizebox{1\linewidth}{!}{
    \begin{tabular}{l|c|ccccc}
        \hline
        \multirow{2}{*}{Method} & Training & \multicolumn{5}{c}{IoU} \\
        & data & @1 & @2 & @3 & @4 & @5 \\
        \hline
        BRS~\cite{jang2019interactive} & SBD & 80.0$^*$ & 85.0$^*$ & 87.0$^*$ & 89.0$^*$ & 91.0$^*$ \\
        f-BRS~\cite{sofiiuk2020f} & SBD & 77.0$^*$ & 83.0$^*$ & 85.0$^*$ & 88.0$^*$ & 90.0$^*$ \\
        EgdeFlow~\cite{hao2021edgeflow} & LC & 80.0$^*$ & 90.0$^*$ & \underline{93.5$^*$} & \underline{94.5$^*$} & \underline{95.0$^*$} \\
        RITM~\cite{sofiiuk2021reviving} & LC & 82.88 & 91.51 & \underline{94.46} & \underline{95.57} & \underline{95.83} \\
        \hline
        Ours & SBD & 93.35 & & & &  \\
        Ours & LC & 93.08 & & & &  \\
        \hline
    \end{tabular}
    }
    \end{center}
    \caption{A quantitative comparison of IS methods on Berkeley. LC denotes LVIS+COCO. “*” means an approximate metric value determined from the IoU plots from the original papers. The results better than ours are \underline{underlined}.}
\label{tab:comparison-berkeley}
\end{table}

\begin{table}[h!]
    \begin{center}
    \tabcolsep=0.1cm
    \resizebox{1\linewidth}{!}{
    \begin{tabular}{l|c|ccccc}
        \hline
        \multirow{2}{*}{Method} & Training & \multicolumn{5}{c}{IoU} \\
        & data & @1 & @2 & @3 & @4 & @5 \\
        \hline
        BRS~\cite{jang2019interactive} & SBD & 72.0$^*$ & 80.0$^*$ & 85.0$^*$ & 86.0$^*$ & 86.0$^*$ \\
        f-BRS~\cite{sofiiuk2020f} & SBD & 71.0$^*$ & 79.0$^*$ & 79.0$^*$ & 82.0$^*$ & 83.0$^*$ \\
        EgdeFlow~\cite{hao2021edgeflow} & LC & 74.0$^*$ & 83.0$^*$ & 86.0$^*$ & \underline{86.05} & \underline{89.0$^*$} \\
        RITM~\cite{sofiiuk2021reviving} & LC & 73.37 & 82.51 & \underline{86.45} & \underline{88.46} & \underline{89.62} \\
        \hline
        Ours & SBD & 85.44 & & & &  \\
        Ours & LC & 86.05 & & & &  \\
        \hline
    \end{tabular}
    }
    \end{center}
    \caption{A quantitative comparison of IS methods on DAVIS. LC denotes LVIS+COCO. “*” means an approximate metric value determined from the IoU plots from the original papers. The results better than ours are \underline{underlined}.}
\label{tab:comparison-davis}
\end{table}

Note, that our method has the same backbone and interactive branch as RITM, so their computational efficiency and inference speed are on par. However, RITM needs more inference rounds to achieve the same segmentation quality. 

\begin{figure}[h!]
    \centering
    \includegraphics[width=0.8\linewidth]{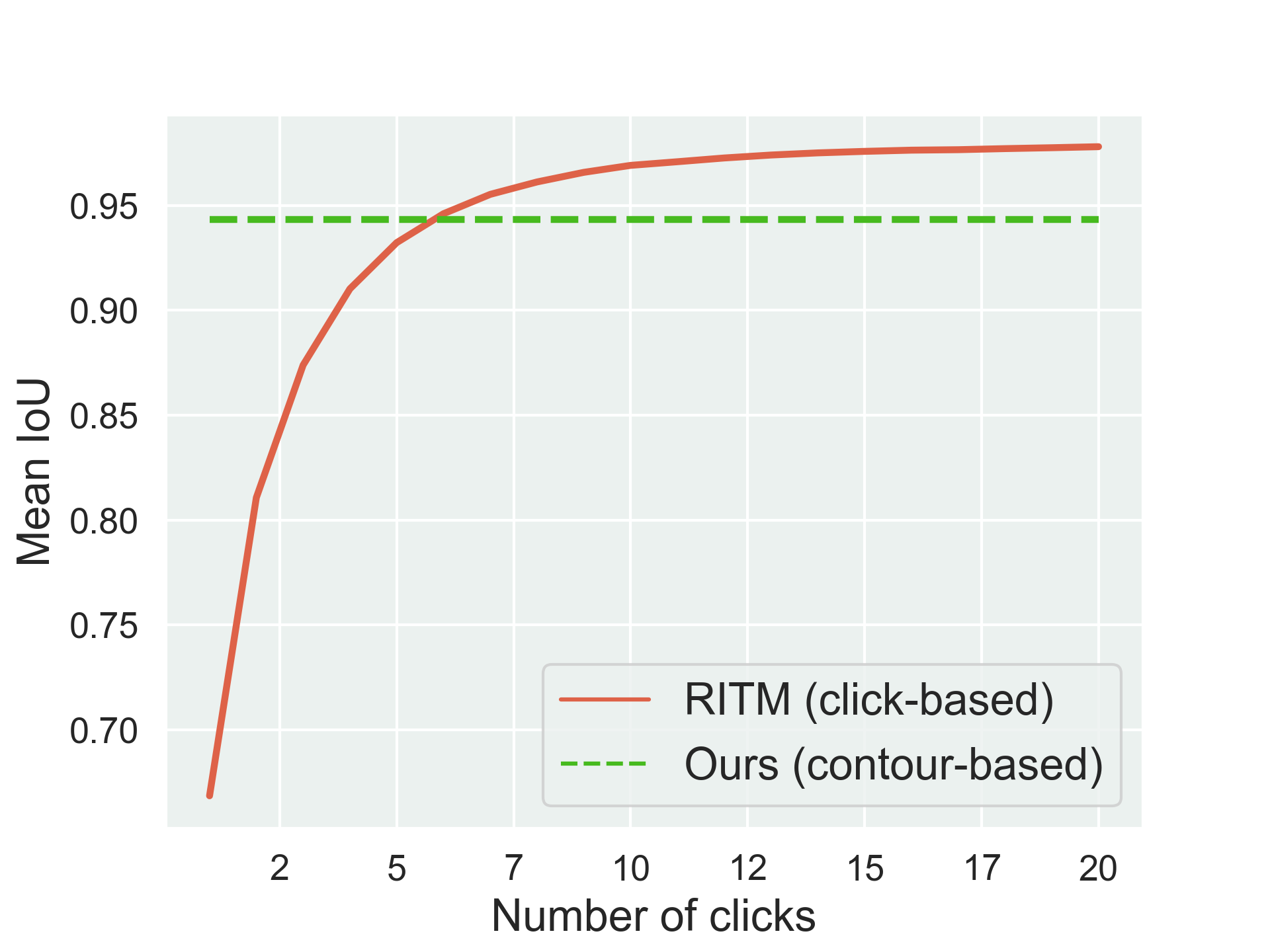}
    \captionof{figure}{IoU of RITM (per click) and IoU of our method (a single contour) on \dset.}
    \label{fig:ritm-quantitative}
\end{figure}

\begin{figure}[h!]
    \centering
    \includegraphics[width=0.8\linewidth]{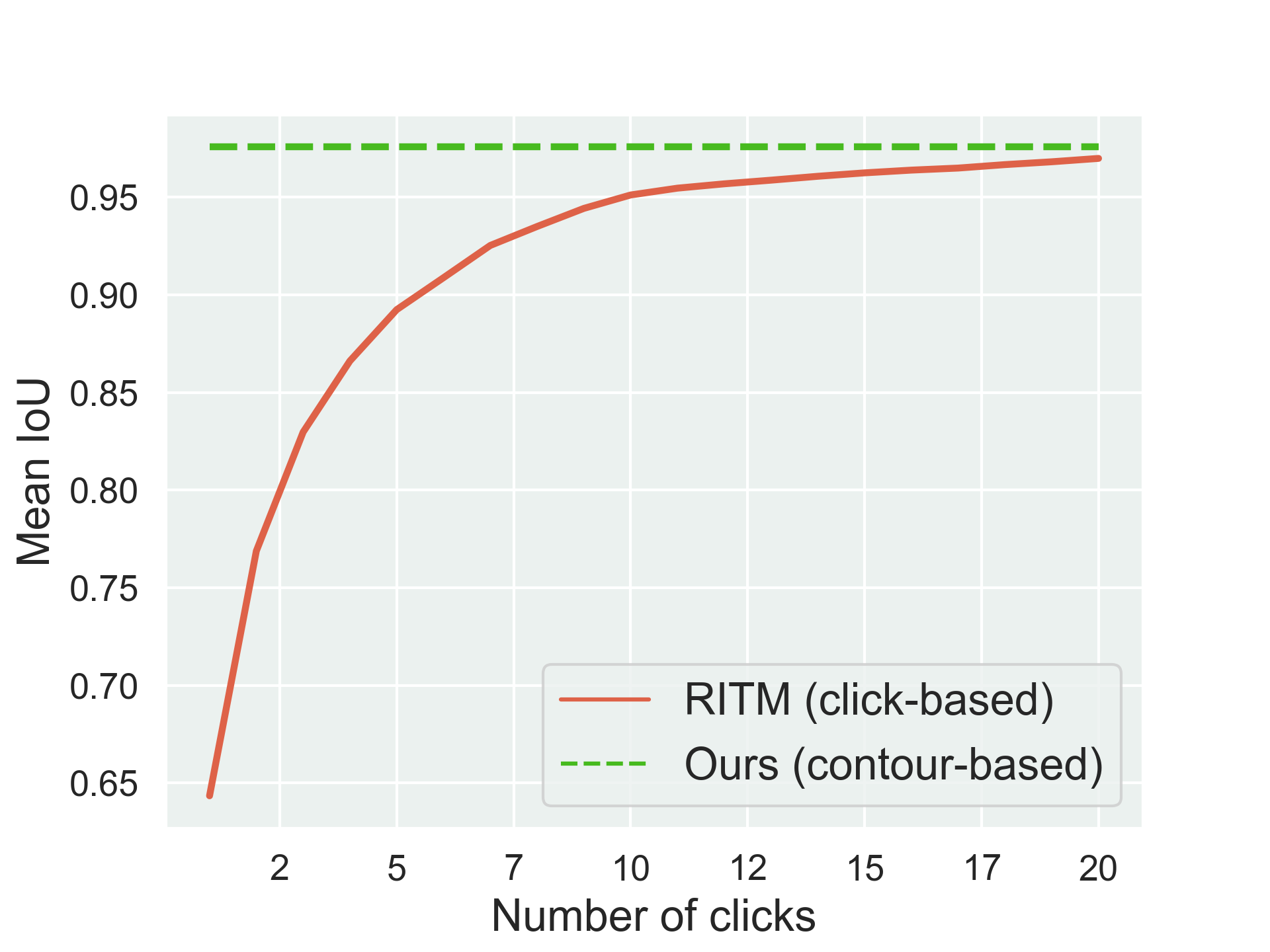}
    \captionof{figure}{IoU of RITM (per click) and IoU of our method (a single contour) on \dsetgroup.}
    \label{fig:ritm-quantitative-groups}
\end{figure}

We compare an IoU per click for RITM with an IoU of our method achieved with a single contour on \dset{}~(Fig.~\ref{fig:ritm-quantitative}) and \dsetgroup{}~(Fig.~\ref{fig:ritm-quantitative-groups}). For \dset, a single contour is as effective as 5 clicks on average. Apparently, for hard segmentation cases present in \dsetgroup, using contours is even more beneficial than using clicks: one contour is equivalent to 20(!) clicks in terms of IoU. A qualitative comparison is presented in Fig.~\ref{fig:ritm-qualitative}.

\subsection{Ablation Study}

\paragraph{Backbones.}
We use the same backbone as RITM~\cite{sofiiuk2021reviving} to guarantee our method can be fairly compared with the previous state-of-the-art. Our experiments with HRNet18s, HRNet32, and HRNet48 reveal that a model complexity does not affect accuracy much (Tab.~\ref{tab:ablation-backbone}), so we opt for the efficient yet accurate HRNet18.

\begin{table}[h!]
    \begin{center}
    \begin{tabular}{c|cccc}
        \hline
        Backbone & GrabCut & Berkeley & DAVIS & SBD \\
        \hline
        HRNet18 & 95.14 & 91.16 & 83.66 & \textbf{87.52} \\
        HRNet18s & 94.28 & \textbf{91.48} & \textbf{83.78} & 86.85 \\
        HRNet32 & \textbf{95.22} & 90.94 & 83.17 & 86.94 \\
        HRNet48 & 94.84 & 90.87 & 82.70 & 87.39 \\
        \hline
    \end{tabular}
    \end{center}
    \caption{Ablation study of backbones. We train our models on LVIS+COCO and represent contours as filled masks. The best results are \textbf{bold}.}
\label{tab:ablation-backbone}
\end{table}

\begin{table*}[h!]
    \begin{center}
    \tabcolsep=0.05cm
    \resizebox{0.925\linewidth}{!}{
    \begin{tabular}{ccccc}
        Ground & RITM, & RITM, & RITM, & Ours,\\
        truth & 1st click & 2nd click & 3rd click & a single contour\\
        \includegraphics[width=90pt]{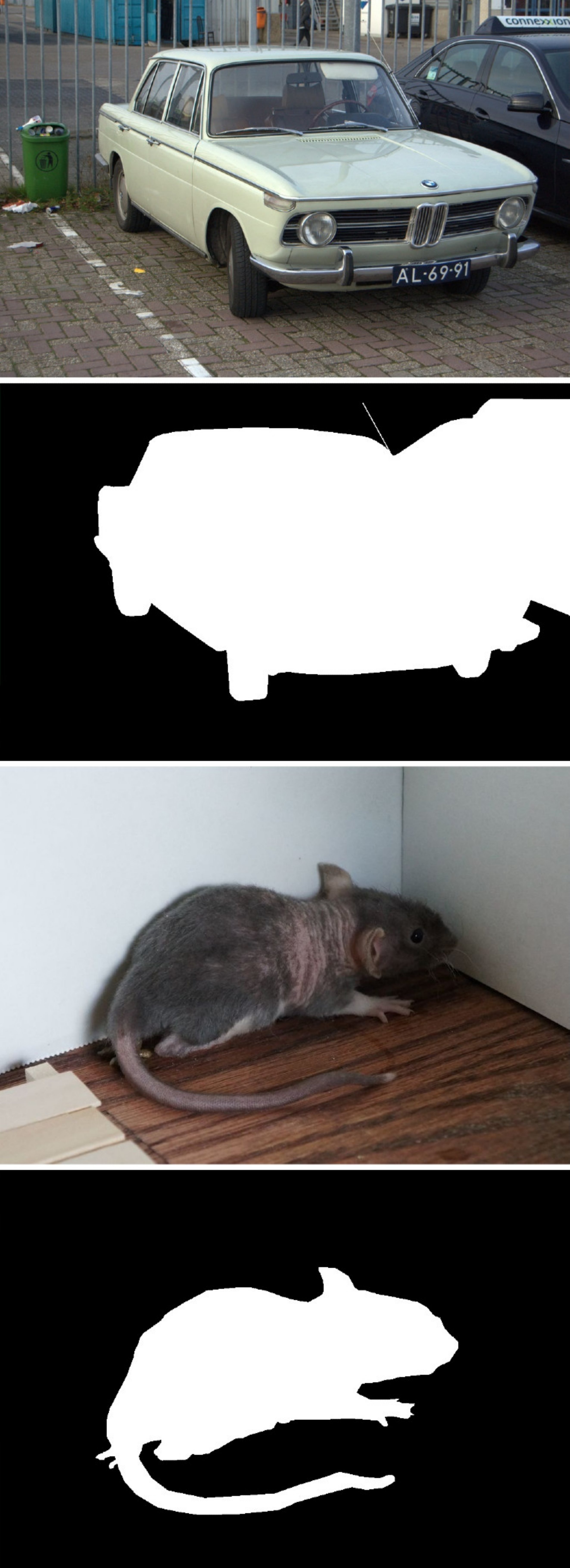} &
        \includegraphics[width=90pt]{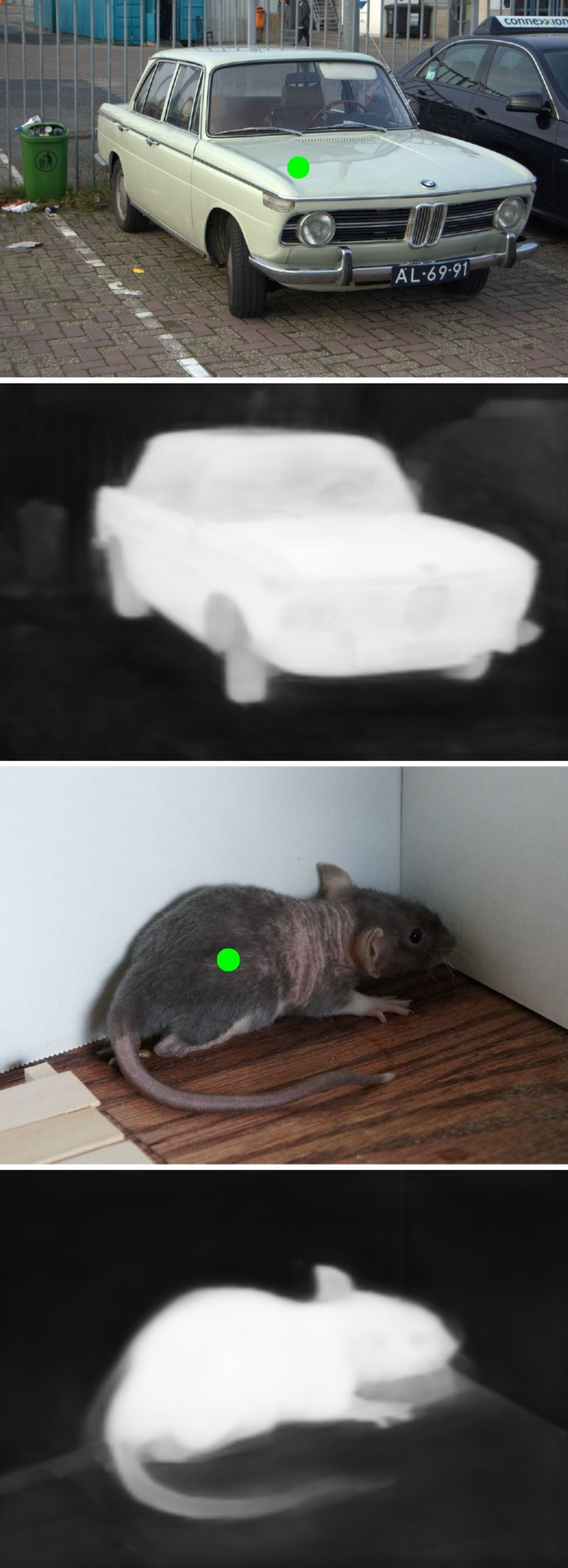} &
        \includegraphics[width=90pt]{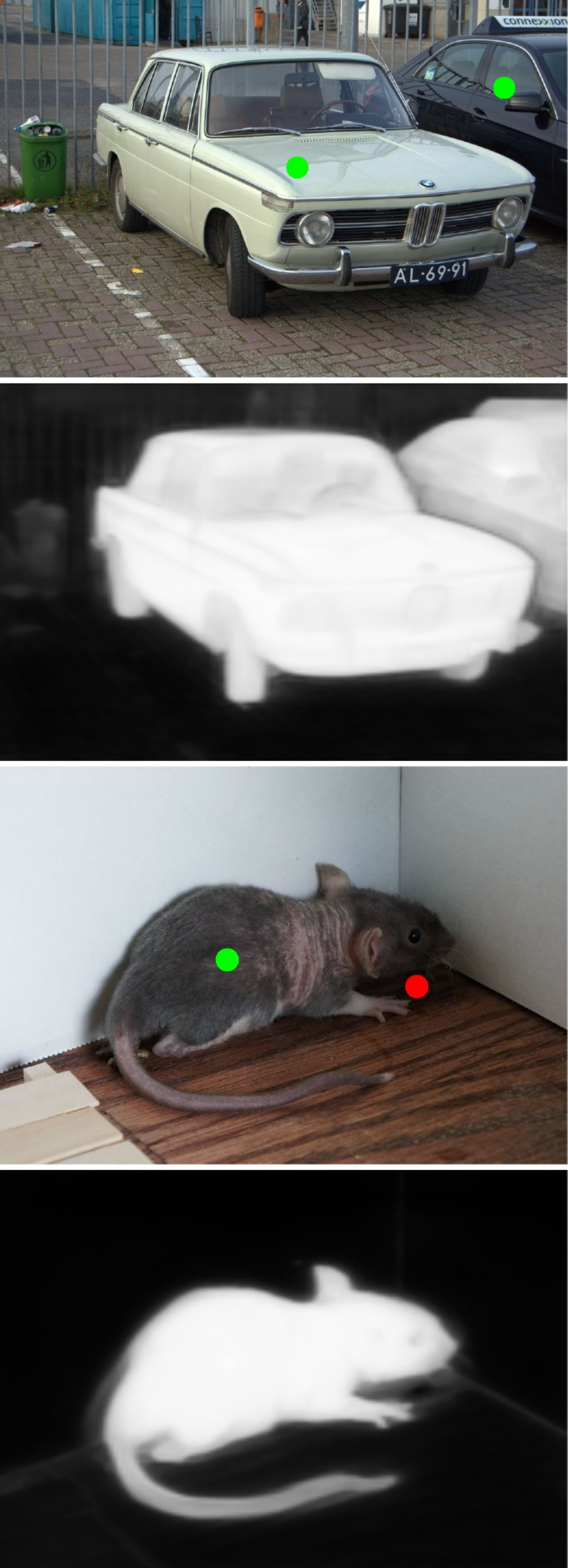} &
        \includegraphics[width=90pt]{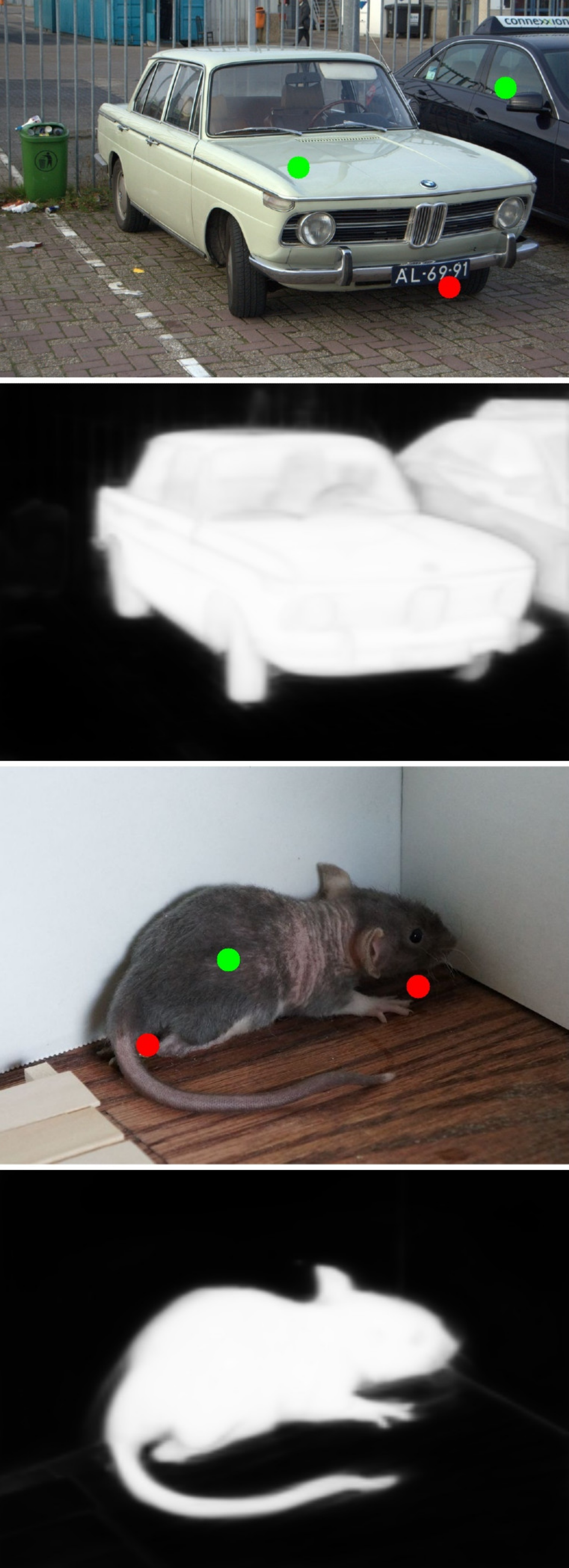} &
        \includegraphics[width=90pt]{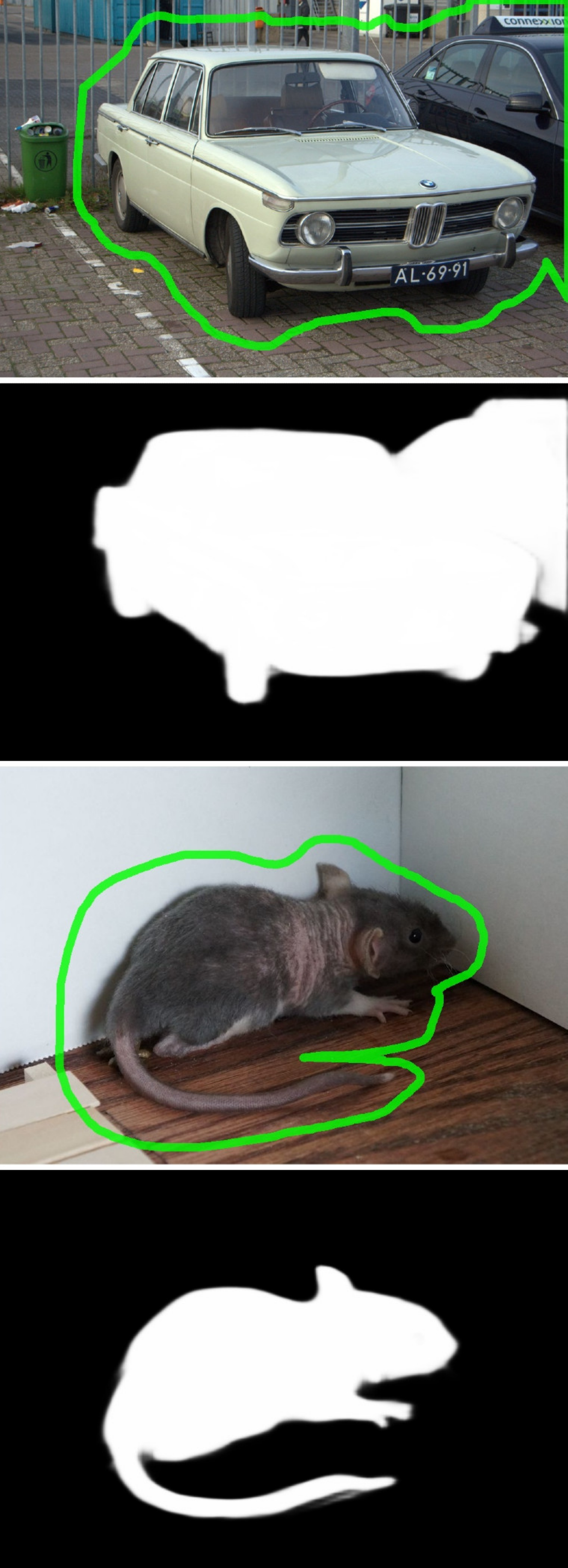} \\
    \end{tabular}
    }
    \end{center}
    \captionsetup{labelformat=empty}
    \captionof{figure}{
        Figure 11: Randomly generated user interactions and corresponding predictions obtained with RITM and our method.
    }
    \label{fig:ritm-qualitative}
\end{table*}

\paragraph{Contours encoding.}

We compare filled contour masks with contours as lines of varying width. According to the Tab.~\ref{tab:ablation-encoding}, a width of 2\% of the length of the shorter side of an image provides the best results among all contour representations in the form of lines. However, they are still inferior to the results obtained with filled contour masks. Respectively, we use “filled” contours in all other experiments, as they facilitate more accurate predictions on all benchmarks. 

\begin{table}[h!]
    \begin{center}
    \begin{tabular}{c|cccc}
        \hline
        Contours & \multirow{2}{*}{GrabCut} & \multirow{2}{*}{Berkeley} & \multirow{2}{*}{DAVIS} & \multirow{2}{*}{SBD} \\
        encoding & & & & \\
        \hline
        Filled & \textbf{95.14} & \textbf{91.16} & \textbf{83.66} & \textbf{87.52} \\
        Line, 0.005 & 25.48 & 16.44 & 25.58 & 21.06 \\
        Line, 0.01 & 94.55 & 90.99 & 81.98 & 87.02 \\
        Line, 0.02 & 94.68 & 91.13 & 82.44 & 87.07 \\
        Line, 0.05 & 94.08 & 89.94 & 81.71 & 86.82 \\
        Line, 0.1 & 94.92 & 89.63 & 80.18 & 86.65 \\
        \hline
    \end{tabular}
    \end{center}
    \caption{Ablation study of contours encoding. We train our models on LVIS+COCO and employ HRNet18 as a backbone. “Line, $w$” means that the contour is represented as a line of a width $=(w \times \text{the length of a shorter image side})$. The best results are \textbf{bold}.}
\label{tab:ablation-encoding}
\end{table}

\paragraph{Training datasets.}

We measure the performance gain from using additional data sources. We leverage the OpenImages data in two different ways. First, we simply combine it with LVIS+COCO for training, following the same training procedure as for LVIS+COCO only. Alternatively, we use a part of the OpenImages data for fine-tuning. To compose a fine-tuning set, we utilize our best model (HRNet18-based, trained on LVIS+COCO with contours represented as filled masks). For each image from OpenImages test split, we generate a random contour and pass it through the model. If a predicted mask has an IoU\textgreater97 with a ground truth mask, we save a sample consisting of an image, a corresponding ground truth mask, and the generated contour. This way, we obtain a set of 2533 contours for 2253 images. 

\begin{table}[h!]
    \begin{center}
    \small
    \begin{tabular}{cc|cccc}
        \hline
        Training & Fine- & \multirow{2}{*}{GrabCut} & \multirow{2}{*}{Berkeley} & \multirow{2}{*}{DAVIS} & \multirow{2}{*}{SBD} \\
        data & tuning & & & & \\
        \hline
        LC & - & 95.14 & 91.16 & 83.66 & 87.52 \\
        LC+OI & - & 95.14 & 92.31 & 83.61 & 87.16 \\
        LC & + & \textbf{96.32} & \textbf{93.08} & \textbf{86.05} & \textbf{87.84} \\
        \hline
    \end{tabular}
    \end{center}
    \caption{Ablation study of training data. We employ HRNet18 as a backbone, and represent contours as filled masks. LC stands for LVIS+COCO, OI means OpenImages. The best results are \textbf{bold}.}
\label{tab:ablation-training-data}
\end{table}

According to Tab.~\ref{tab:ablation-training-data}, training on OpenImages does not lead to a performance gain: the results are better for Berkeley and worse for DAVIS and SBD, compared to training only on LVIS+COCO. However, fine-tuning improves the quality for all test datasets. Evidently, it is more profitable to fine-tune on a carefully selected minor subset of OpenImages, than to use the entire test+validation split for training.

\section{Conclusion}
We presented a novel contour-based interactive segmentation method. We tested our approach on standard benchmarks against click-based methods and showed that a single contour provides the same accuracy as several clicks. Moreover, we introduced a novel \dset{} containing human-annotated contours for common objects in the wild, and \dsetgroup{}, featuring difficult segmentation cases. We empirically proved that our contour-based approach has an even greater advantage over click-based methods on challenging data. Overall, we demonstrated that contours could reduce the required number of interactions and significantly simplify image editing and labeling.

\vfill\clearpage

\bibliographystyle{named}
\bibliography{ijcai23}

\appendix

\section{Contour Generation}

In the main paper, we provide a brief overview of our contour simulation procedure. Here, we would like to discuss it in detail, providing values of hyperparameters and explicating the motivation behind design choices. Given a ground truth segmentation mask, we generate contours as follows:

\begin{enumerate}
    \item First, we fill all holes in the mask. 

    \item After that, we randomly select either dilation or erosion. For a chosen morphological transform, we sample: $d_{dilation} \sim Uniform(0.03, 0.06), \ d_{erosion} \sim Uniform(0.02, 0.04)$ and multiply this value by an image diagonal length to obtain a kernel size. Hence, the transformed mask does not stretch or shrink too much, yet close-by objects might merge, so that the contour encloses a group of objects.

    \item A contour should not outline two distant parts of an object or even two different objects, so we do not consider disconnected areas of the transformed mask. Accordingly, we search for connected components and select the one with the largest area. 

    \item Furthermore, we distort the mask via an elastic transform. The parameters of the elastic transform are randomly generated: a shift $d_{affine} \sim Uniform(0.4 , 0.6)$, a standard deviation for GaussianBlur $d_{sigma} \sim Uniform(0.5, 0.75)$, and a normalizing coefficient $d_{alpha} \sim Uniform(0.8, 1.2)$, which controls the intensity of the deformation. All the parameters are then multiplied by the size of the object. The elastic transform might divide the mask into several disconnected parts, so we select the largest connected component yet again.

    \item We smooth the mask via GaussianBlur. We sample $d_{size} \sim Uniform(0.008, 0.06)$ and multiply it by the object size to obtain a random kernel size.

    \item Next transformation is a random scaling. We assume that objects of a simple shape might be outlined rather coarsely, while complex shapes require a more thoughtful approach. Guided by this observation, we define a ratio $r$ reflecting the “complexity” of the object shape: it is calculated as the area of the current mask by the area of its convex hull. If $r < 0.6$, we assume an object has a complex, non-convex shape. In this case, we cannot apply severe augmentations to the mask, since the distorted mask would match the object badly. If $r \ge 0.6$, an object seems to be “almost convex”, so intense augmentations would not affect its shape so dramatically. Accordingly, we randomly sample a scaling factor $d_{scale} \sim Uniform(0.9, 1.1)$ for complex, “non-convex” objects and $d_{scale} \sim Uniform(0.75, 1.2)$ for less complex, “almost convex” objects.

    \item Then, we select a proper shift based on the size of an object. Particularly, we consider bounding boxes enclosing the transformed and the ground truth masks. These bounding boxes are parametrized with min-max coordinates along coordinate axes, so the transformed bounding box is represented as $(x_0^{tr}, y_0^{tr}, x_1^{tr}, y_1^{tr})$ and the ground truth bounding box is represented as $(x_0^{gt}, y_0^{gt}, x_1^{gt}, y_1^{gt})$. We compute axis-wise Manhattan distances: 
    \begin{equation}
    \begin{aligned}
        d_x &= \min \{ | x_0^{tr} - x_0^{gt} |, | x_1^{tr} - x_1^{gt} | \} \\
        d_y &= \min \{ | y_0^{tr} - y_0^{gt} |, | y_1^{tr} - y_1^{gt} | \}, \\
    \end{aligned}
    \end{equation}
    and multiply $d_x$ and $d_y$ by 2 for the “almost convex” objects.
    The distances are clipped so that they do not exceed $0.5$ of the size of the transformed bounding box along the corresponding axes:
    \begin{equation}
    \begin{aligned}
        d_x &= \min\{ d_x,  0.5 (x_1^{tr} - x_0^{tr}) \} \\
        d_y &= \min\{ d_y,  0.5 (y_1^{tr} - y_0^{tr}) \} \\
    \end{aligned}
    \end{equation}
    Finally, we randomly select an integer value $d_{shift_x} \sim Uniform(\{-d_x, \dots, 0, \dots,  d_x\})$ and an integer value $d_{shift_y} \sim Uniform(\{-d_y, \dots, 0, \dots,  d_y\})$ as shifts along the $x$-axis and $y$-axis, respectively.  
\end{enumerate}

\section{Data Labeling}

\subsection{Annotators}
\label{ssec:annotators}

We contracted a company specializing in data markup for labeling our data. The company hired annotators to perform the task, and was in response for training, management, and quality control.

We decomposed the labeling task into two subtasks. The first subtask implies creating instance segmentation masks for the given images. For the test subsets of the standard benchmarks, we use instance segmentation masks already present in these datasets. The second subtask is to outline previously segmented instances with contours.

\subsection{Annotator Guide}
\label{ssec:annotator-guide}

We provided our contractor with the following guide.

\paragraph{Instance labeling.} 

\begin{enumerate}
    \item The annotator labels the images with object classes and transfers back instance segmentation masks as PNG images of the same resolution as original RGB images; each object in a mask should have a unique color;
    
    \item On each image, at least one object of interest should be labeled. The choice of object is up to the annotator. 
    
    \item Since either an individual object, a group of close-by objects, or a subgroup can be a subject of interest, the annotators label 50\% of instance masks as groups and 50\% as individual objects. 

\end{enumerate}

\paragraph{Contour labeling}

\begin{enumerate}
    \item The annotator outline each segmented instance with a contour. 
    
    \item There should be no breaks in a contour; however, its start may not coincide with its end. 
    
    \item The contours should not be extremely accurate, but drawn naturally. Nevertheless, the correspondence between instances and contours should be clear and unambiguous.
    
    \item If an object is overlapped, the annotators might use negative contours to outline an overlapping object. Alternatively, multiple positive contours might be used if an object contains several disconnected parts.
    
\end{enumerate}

\section{Dataset Structure}

\subsection{File Structure}

The root directory has two subdirectories with images and masks, respectively, and a single JSON file containing an annotation. 

\begin{lstlisting}[language=Python]
    dataset/
        contours.json
        images/
            0000001.jpg
            0000002.jpg
            0000003.jpg
            ...
            0015000.jpg
        masks/
            0000001_01.png
            0000001_02.png
            0000001_03.png
            0000002_01.png
            0000002_02.png
            0000003_01.png
            0000003_02.png
            ...
            0015000_01.png
            0015000_02.png
\end{lstlisting}

Images are saved as JPG files with names in the format [IMAGE\_ID].jpg. For a single image IMAGE\_ID, there might be multiple instance masks labeled by different annotators. The corresponding instance masks are saved as PNG files named [IMAGE\_ID]\_[ANNOTATION\_NUMBER].png, where annotation numbers (starting from 01) indicate different versions of instance mask labeling. 

\subsection{Annotation Format}

The annotations are saved as a JSON file. This file contains a dictionary, where keys are image identifiers, and values are lists. Each item in a list is a dictionary describing an annotation provided by a single annotator (as we collected multiple annotations per image). This dictionary features two keys: \textit{pos\_contours} and \textit{neg\_contours}. For each key, there is a list of contours of the corresponding type. 

A contour is actually a polygon, represented as an ordered list of coordinates of its vertices in an image. The coordinates are given in normalized image coordinates varying from 0 to 1 along each image side.

\subsection{Implementation Details}

\paragraph{Software \& hardware.}
Training on LVIS+COCO takes 81 hours on a workstation equipped with two Tesla P40 GPUs. Inferring a single user interaction (an RGB image along with previous masks) takes about 80 milliseconds on a single RTX 3090 GPU. All our models are implemented with PyTorch~\cite{paszke2019pytorch}. We rely on the original implementation of HRNet\footnote{\resizebox{0.93\linewidth}{!}{{\url{https://github.com/HRNet/HRNet-Image-Classification}}}} and use the pre-trained ImageNet~\cite{deng2009imagenet} weights provided in the official repository.

\end{document}